\documentclass{article}

\PassOptionsToPackage{numbers, compress}{natbib}
 \usepackage[preprint]{neurips_2026}

\usepackage[utf8]{inputenc} % allow utf-8 input
\usepackage[T1]{fontenc}    % use 8-bit T1 fonts
\usepackage{url}            % simple URL typesetting
\usepackage{booktabs}       % professional-quality tables

\usepackage{amsfonts}       % blackboard math symbols
\usepackage{nicefrac}       % compact symbols for 1/2, etc.
\usepackage{microtype}      % microtypography
\usepackage{xcolor}         % colors
\usepackage{graphicx}
\usepackage{enumitem}
\usepackage{amsmath}
\usepackage{wrapfig}        
\usepackage{gradient-text}
\usepackage{setspace}
\usepackage{multirow}
\usepackage{algorithm}    
\usepackage{algpseudocode}
\usepackage{colortbl}
\usepackage{soul}
\usepackage{caption}
\usepackage{makecell}
\usepackage[normalem]{ulem}

\definecolor{blue}{rgb}{0.21,0.49,0.74}
\usepackage[colorlinks=true, linkcolor=blue, citecolor=blue, urlcolor=blue]{hyperref}
\usepackage[capitalise]{cleveref}       % for \cref references
\definecolor{customColor}{HTML}{FFF3E9}
\definecolor{mdriveorange}{HTML}{F98327}
\definecolor{mdrivepurple}{HTML}{642991}
\definecolor{mdriveblue}{HTML}{23538c}

\title{%
  \raisebox{-.18\height}{\includegraphics[width=0.0853\textwidth]{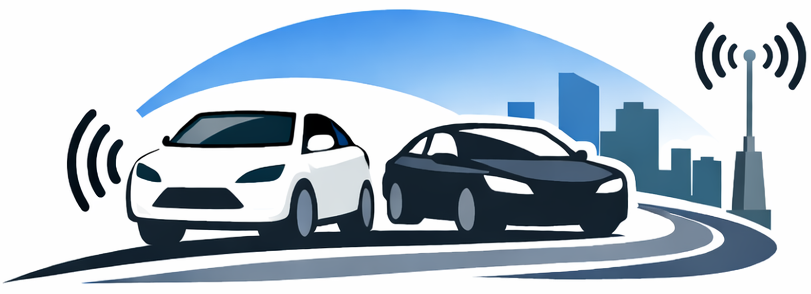}} \hspace{0.0001em}%
  \textit{MDrive}: Benchmarking Closed-Loop Cooperative Driving for End-to-End Multi-agent Systems
}

\author{
Marco Coscoy\thanks{Equal contributions. $^\ddagger$Project leads. Emails: \texttt{\{mcoscoy, zeweizhou, sethzhao506\}@ucla.edu}.}\quad
Zewei Zhou\textsuperscript{*,\,$\ddagger$}\enspace
Seth Z. Zhao\textsuperscript{*,\,$\ddagger$}\enspace
Henry Wei\enspace
Angela Magtoto\enspace
Johnson Liu
\AND
Rui Song\thanks{Corresponding author. Email: \texttt{rruisong@ucla.edu}.}\quad
Walter Zimmer\quad
Zhiyu Huang\quad
Chen Tang\quad
Bolei Zhou\quad
Jiaqi Ma \\ [0.1cm] 
University of California, Los Angeles
\\ [0.1cm]
\small \tt{\href{https://mdrive-challenge.github.io/}{https://mdrive-challenge.github.io/}}
}

\begin{document}

\maketitle

\begin{figure}[tbh]
    \centering
    \includegraphics[width=\linewidth]{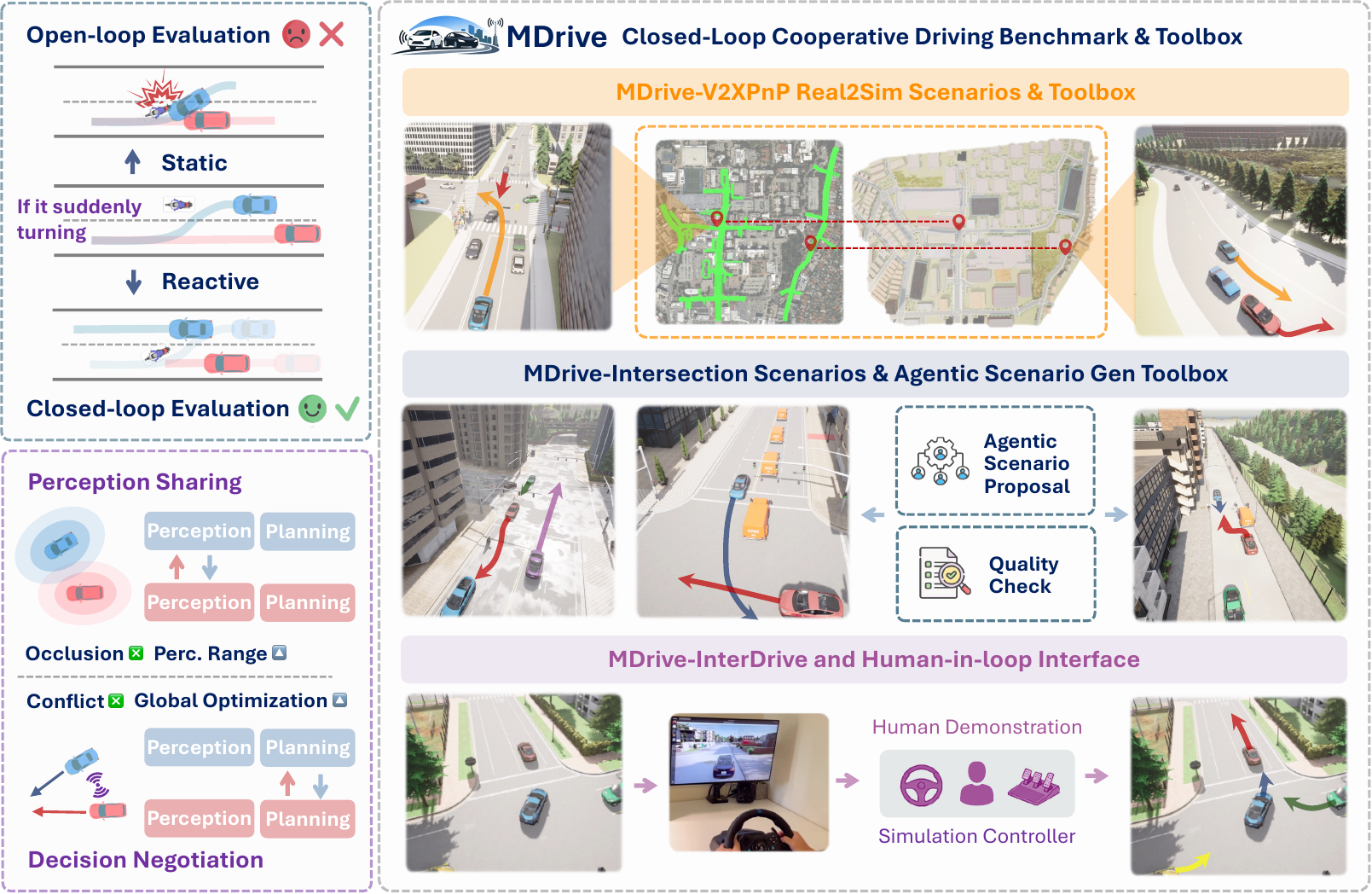}
    % \vspace{0.01cm}
    \caption{\textbf{MDrive Overview.} MDrive is a closed-loop cooperative driving benchmark. To the best of our knowledge, it is the first benchmark to systematically evaluate the real benefits of multi-agent cooperation, including the perception sharing and decision negotiation. MDrive benchmarks include three scenario buckets: MDrive-V2XPnP Real2Sim scenario reconstructed from real-world V2X driving logs, MDrive-Intersection generated by the agentic scenario proposer, and MDrive-InterDrive. The Real2Sim, agentic scenario generation, and human-in-loop toolbox of MDrive is released.}
    \label{fig:mdrive_hero}
\end{figure}

\begin{abstract}
Vehicle-to-Everything (V2X) communication has emerged as a promising paradigm for autonomous driving, enabling connected agents to share complementary perception information and negotiate with each other to benefit planning. Existing V2X benchmarks, however, fall short in two ways: (\emph{i}) open-loop evaluations fail to capture the inherently closed-loop nature of driving, leading to evaluation gaps, and (\emph{ii}) current closed-loop evaluations lack behavioral and interactive diversity to reflect real-world driving. Thus, it is still unclear the \emph{extent of benefits} of multi-agent systems for closed-loop driving. In this paper, we introduce \textbf{MDrive}, a closed-loop cooperative driving benchmark comprising 225 scenarios grounded in both NHTSA pre-crash typologies and real-world V2X datasets. Our benchmark results demonstrate that multi-agent systems are generally better than single-agent counterparts. However, current multi-agent systems still face two important challenges: (\emph{i}) perception sharing enhances perceptions, but doesn't always translate to better planning; (\emph{ii}) negotiation improves planning performance but harms it in complex and dense traffic scenarios. MDrive further provides an open-source toolbox for scenario generation, Real2Sim conversion, and human-in-the-loop simulation. Together, MDrive establishes a reproducible foundation for evaluating and improving the generalization and robustness of cooperative driving systems.
\end{abstract}

\section{Introduction}
Vehicle-to-everything (V2X) communication has emerged as a promising paradigm for multi-agent driving systems~\cite{zhong2025cooptrack,xu2022opv2v,li2022v2x,song2024collaborative}, extending autonomous driving from isolated single-agent operation to cooperative perception and negotiation. As illustrated in \cref{fig:mdrive_hero}, by allowing agents to share perception cues (i.e., perception sharing) and intent or planning information (i.e., decision negotiation), V2X communication can expand the effective perception range, mitigate occlusion~\cite{zhou2024v2xpnp, zimmer2024tumtrafv2x,v2xrealo, zhao2025quantv2x, zhao2024coopre, song2024collaborative, zhou2025turbotrain}, and enable more proactive planning beyond passive single-agent decision making~\cite{yu2025end,lei2025cooperrisk}. However, most previous work has focused on demonstrating gains in multi-agent open-loop driving metrics~\cite{liu2025mmcooper,zhang2025mic,lei2025risk}, which fails to reflect actual performance gains in the inherently closed-loop task of driving \cite{zhao2026bridgesim}. This motivates a fundamental question: \textbf{\emph{To what extent does multi-agent cooperation benefit closed-loop driving, and where do these gains come from?}} 

To answer this question, we argue that existing evaluation protocols are insufficient for benchmarking multi-agent cooperative driving in closed-loop settings. We identify two key gaps.
First, prevailing open-loop protocols for cooperative perception and prediction~\cite{zhou2024v2xpnp, zimmer2024tumtrafv2x,v2xrealo, zhao2025quantv2x, zhao2024coopre, song2024collaborative, zhou2025turbotrain} evaluate static outputs in non-reactive environments. Those protocols do not test whether improved perception or prediction leads to better closed-loop driving.
Therefore, such protocols miss core closed-loop abilities, such as failure recovery and adaptation to interactive agents, and often correlate weakly with closed-loop performance~\cite{zhao2026bridgesim, ol_cl_survey, li2024ego, jia2024bench2drive}, as also shown in our experiments in Section~\ref{subsec:ol_cl}.
Second, existing closed-loop multi-agent benchmarks still lack sufficient realism and diversity, as illustrated in Table~\ref{tab:benchmark_comparison}. 
InterDrive~\cite{colmdriver} is the most relevant existing benchmark to our setting, but its scenarios are primarily geared toward negotiation-style interactions among a small number of cooperating agents. Its scenarios do not model background traffic and cover only limited road layouts. 
This simplified setting remains far from realistic urban driving, making it difficult to evaluate the robustness and generalizability of cooperative driving systems \cite{colmdriver, cui2026coopreflect}.

To close these gaps, we introduce \textbf{MDrive}, a closed-loop cooperative driving benchmark for systematically evaluating the real benefits of multi-agent cooperation.
MDrive comprises 225 multi-agent scenarios drawn from two sources: (\emph{i}) NHTSA pre-crash typologies~\cite{najm2007pre} and curated long-tail interaction categories such as work zones, roundabouts, unprotected left turns, and zipper merges, generated through an agentic scenario synthesis pipeline; and (\emph{ii}) 
Reactive scenarios converted from real-world V2X driving logs through a Real-to-Simulation (Real2Sim) pipeline.
These scenarios expose models to realistic background traffic that is difficult to obtain from purely synthetic benchmarks.
Alongside the MDrive benchmark, we release \textbf{MDrive-Toolbox}, an open-source pipeline for agentic scenario generation, Real2Sim conversion, and a human-in-the-loop interface, lowering the barrier for the community to build, extend, and customize closed-loop cooperative driving benchmarks.

With the aforementioned scenario sources and interfaces, we first examine whether multi-agent systems outperform single-agent systems. We find a clear overall advantage for multi-agent systems, with both perception-sharing and negotiation-based paradigms outperforming their single-agent counterparts on average. However, benchmarking on MDrive reveals that these two multi-agent paradigms still suffer from the following limitations. First, for cooperative perception, our findings challenge the prevailing assumption that better perception always leads to better planning. Specifically, we observe that even though perception sharing can compensate for limited observability, it does not guarantee safe decision-making in ambiguous situations. This is related to recent findings~\cite{LEAD, gerstenecker2026fail2drive, zhao2026bridgesim} that current open-loop training through offline demonstrations is still insufficient to ensure generalizable closed-loop planning capabilities. On the other hand, we find that the benefits from the negotiation-based method are limited to particular scenarios with sparse and simple traffic, as current language-based agent often make hallucinated decisions in complex and dense traffic scenarios with rich urban interaction (see \cref{subsec:rq2}). We further compare the negotiation method to a real human expert (see Section~\ref{subsec:human_study}) and find that it still exhibits a large behavior gap in decision-making under complex interaction scenarios.

Taken together, MDrive addresses existing evaluation gaps in assessing the closed-loop driving capabilities of multi-agent systems.
Our analysis further reveals the benefits and remaining challenges of multi-agent cooperation in current models. Our key contributions can be summarized as follows: 
 
\begin{enumerate}[leftmargin=1.5em] 
\item  We propose \textbf{MDrive}, a closed-loop cooperative driving benchmark for end-to-end multi-agent systems. 
MDrive contains 225 diverse scenarios with different interactivity levels, covering Real2Sim scenes converted from real-world V2X datasets and agentic-generated scenarios grounded in predefined interaction categories and NHTSA standards.

\item We release \textbf{MDrive-Toolbox}, an open-source toolbox for scalable benchmark construction, integrating agentic scenario generation, Real2Sim conversion, and human-in-the-loop interface.

\item We conduct extensive analysis and demonstrate the benefits of cooperation in planning with perception sharing and decision negotiation, and discuss their failure cases. Our cross-paradigm evaluation further reveals the limitations of open-loop metrics and highlights the necessity of closed-loop evaluation for cooperative driving systems.
\end{enumerate}

\begin{table}[t]
  \vspace{-0.3cm}
  \caption{\textbf{Comparisons with existing end-to-end open-loop and closed-loop driving benchmarks.} Bg. Actors: background actors; Reactive: reactive simulation; Perc. Sharing: perception sharing.}
  \vspace{0.2cm}
  \label{tab:benchmark_comparison}
  \centering
  \footnotesize
  \setlength{\tabcolsep}{3pt}
  \renewcommand{\arraystretch}{1.15} 
  \begin{tabular}{l|cccc|cc|c}
    \toprule
    \textbf{Benchmark} & \textbf{E2E} & \textbf{Closed-loop} & \textbf{Bg. Actors} & \textbf{Reactive} & \textbf{Perc. Sharing} & \textbf{Negotiation} & \textbf{Scenario Gen.} \\
    \midrule
    \rowcolor[HTML]{eaf1f8} \multicolumn{8}{l}{\textit{Single-agent Driving Benchmarks}} \\
    nuPlan~\cite{caesar2021nuplan}   & $\times$ & $\times$ & \checkmark & $\times$ & $\times$ & $\times$ & Log-replay \\
    NAVSIM~\cite{dauner2024navsim, navsimv2}           & \checkmark & $\times$ & \checkmark & $\times$ & $\times$ & $\times$ & Log-replay \\
    Bench2Drive~\cite{jia2024bench2drive}  & \checkmark & \checkmark & \checkmark & \checkmark & $\times$ & $\times$ & Hand-crafted \\
    HUGSIM~\cite{zhou2024hugsim}     & \checkmark & \checkmark & \checkmark & \checkmark & $\times$ & $\times$ & Real-world Data \\
    DriveArena~\cite{yang2024drivearena}     & \checkmark & \checkmark & \checkmark & \checkmark & $\times$ & $\times$ & Real-world Data \\
    Fail2Drive \cite{gerstenecker2026fail2drive} & \checkmark & \checkmark & \checkmark & \checkmark & $\times$ & $\times$ & Hand-crafted \\
    BridgeSim \cite{zhao2026bridgesim}           & \checkmark & \checkmark & \checkmark & \checkmark & $\times$ & $\times$ & Real2Sim \\
    \midrule
    \rowcolor[HTML]{fff3e9} \multicolumn{8}{l}{\textit{Multi-agent Driving Benchmarks}} \\
    RiskMM~\cite{lei2025risk}           & \checkmark & $\times$ & \checkmark & $\times$ & \checkmark & $\times$ & Real-world Data \\
    V2Xverse \cite{codriving}  & \checkmark & \checkmark & \checkmark & \checkmark & \checkmark & $\times$ & Hand-crafted \\
    InterDrive \cite{colmdriver} & \checkmark & \checkmark & $\times$ & $\times$ & $\times$ & \checkmark & Hand-crafted \\
    \textbf{MDrive} \hspace{0.0001em} \raisebox{-.18\height}{\includegraphics[width=0.07\textwidth]{figure/MDrive-Logo.png}} & \checkmark & \checkmark & \checkmark & \checkmark & \checkmark & \checkmark & Real2Sim, Agentic-Gen \\
    \bottomrule
  \end{tabular}
  \vspace{-0.1cm}
\end{table}

\section{Related Work}
\textbf{End-to-end Autonomous Driving.} 
Traditional autonomous driving systems follow a modular paradigm with independent perception \cite{li2024bevformer,lang2019pointpillars}, prediction \cite{shi2023mtr,zhou2023qcnext,zhou2022comprehensive}, and planning \cite{zhou2021reliable,huang2024gen} components. While this structure is easier to design and maintain, it is limited by error accumulation across modules \cite{chen2024end}. In contrast, end-to-end approaches have become the mainstream, which eliminate inter-module error propagation, simplify the system, and enable joint optimization for the final planning task \cite{DrivoR,RAP,zhou2025autovla,zhou2026spanvla}. However, current paradigms primarily focus on single-agent systems and remain constrained by local perception and passive planning \cite{diffusiondrivev2,LEAD}. Long-tail scenarios have become a key focus in autonomous driving \cite{gerstenecker2026fail2drive,waymo2025e2e}. However, the potential of multi-agent systems, i.e., perception sharing and decision negotiation, especially in scenarios where existing systems fail, remains underexplored.

\textbf{Cooperative Driving System.} 
Multi-agent systems facilitate seamless information sharing and cooperative driving between CAVs and intelligent infrastructure, offering a promising paradigm beyond single-agent systems \cite{disconet,li2021learning,Where2comm:22,HEAL,v2xadversial}. Despite this potential, current research remains in cooperative perception \cite{zhao2025coopre,ffnet}, which is the initial stage. How to extend these cooperative benefits to downstream tasks, especially the final planning task, is the core question for the existing work. The V2XPnP \cite{zhou2024v2xpnp} and TurboTrain \cite{zhou2025turbotrain} proposed a spatial-temporal feature fusion framework, benchmark, and training strategy as the foundation for the downstream task. CooperTrack \cite{zhong2025cooptrack} explores the downstream tracking task. Although some emerging frameworks have begun investigating open-loop perception sharing \cite{yu2025end,chiu2025v2v,you2026v2x} and closed-loop negotiation \cite{colmdriver,cui2026coopreflect} for planning, how the driving performance is benefited by V2X with perception sharing and decision negotiation remains an open question.

\textbf{Driving Simulators and Benchmarks.} 
Driving simulators and benchmarks offer efficient, safe, and flexible environments for model development \cite{wu2024recent}, particularly for closed-loop simulations that demand reactive behaviors beyond the scope of static driving logs. As \cref{tab:benchmark_comparison} shows, while traditional high-fidelity simulators like CARLA \cite{carla}, MetaDrive \cite{li2022metadrive}, and DeepDrive \cite{deepdrive} provide configurable environments and physical modeling, they are often constrained by handcrafted assets and limited visual realism. To bridge this gap, recent data-driven and generative frameworks, notably HUGSIM \cite{zhou2024hugsim} and RAD \cite{gao2025rad}, leverage 3D Gaussian Splatting~\cite{zhou2024drivinggaussian,song2025coda,yan2024street,song2026energs,chen2026periodic} for photorealistic, real-time reconstruction. However, within the domain of cooperative driving, the intricacies of multi-agent perception and planning make the CARLA-based platforms the mainstream \cite{xu2021opencda,liu2025towards}. However, there remains a critical void: a lack of complex scenarios for evaluating perception sharing and decision negotiation, and the absence of a comprehensive closed-loop benchmark for cooperative driving.

\begin{figure}[t]
    \centering
    \includegraphics[width=\linewidth]{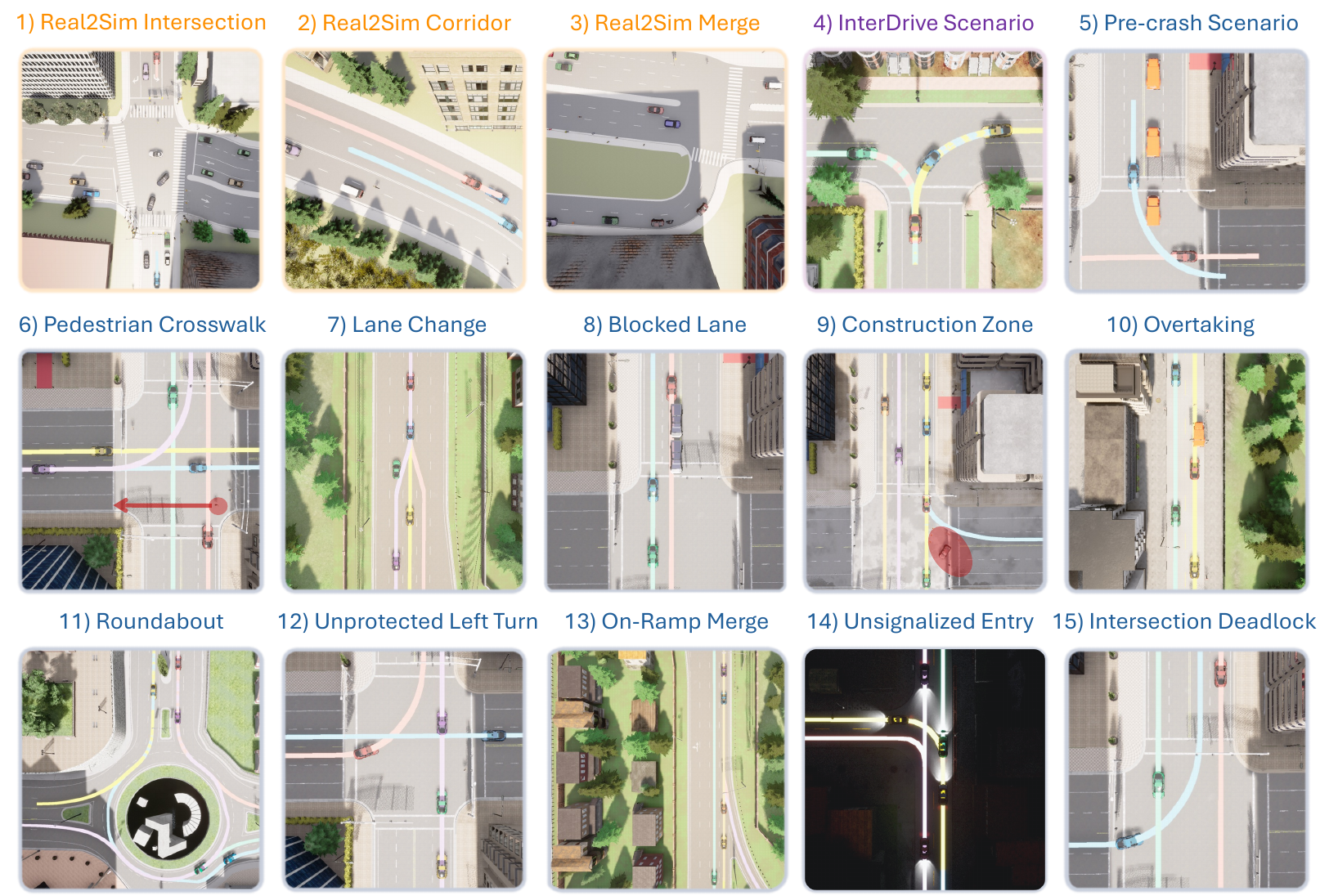}
    \vspace{0.02cm}
    \caption{\textbf{MDrive Representative Scenarios.} The colorful lines in each scenario represent the target routes as the local planning reference for CAVs. \textcolor{mdriveorange}{(1-3)} showcase \textit{\textcolor{mdriveorange}{MDrive-V2XPnP}} Real2Sim scenarios constructed from real-world driving logs, capturing authentic scenario layouts and agent behaviors. \textcolor{mdrivepurple}{(4)} illustrates an \textit{\textcolor{mdrivepurple}{MDrive-Interdrive}} scenario, serving as an in-domain reference without background actors. The remaining cases \textcolor{mdriveblue}{(5-15)} highlight \textit{\textcolor{mdriveblue}{MDrive-Interaction}}, which covers three interactivity levels: \textit{Static Avoidance} (e.g., \textcolor{mdriveblue}{(7)} construction zone, \textcolor{mdriveblue}{(8)} blocked lane, and \textcolor{mdriveblue}{(10)} overtaking), \textit{Dynamic Avoidance} (e.g., \textcolor{mdriveblue}{(6)} pedestrian crosswalk), and \textit{Dynamic Coordination} (e.g., \textcolor{mdriveblue}{(5,9,11-15)}). MDrive also covers different weather and light conditions, such as the night \textcolor{mdriveblue}{(11)}, and a rainy day in \textcolor{mdriveblue}{(7)}.}
    \label{fig:mdrive_representative_scenarios}
    \vspace{-0.2cm}
\end{figure}

\section{MDrive Benchmark}
 MDrive is a closed-loop cooperative driving benchmark built on CARLA~\cite{carla} simulator, which aims to provide broad, structured coverage of multi-agent scenarios to bridge the existing evaluation gap. MDrive includes 225 scenarios and combines three complementary scenario buckets: 1) \textit{MDrive-Interdrive}: in-domain performance of existing models without background actors; 2) \textit{MDrive-Interaction}: interaction scenarios that enrich background actors interactivity and diversity; 3) \textit{MDrive-V2XPnP} Real2Sim scenarios for testing models under real-world scenario layout and behaviors. 
 % The resulting benchmark contains 225 closed-loop scenarios.

\subsection{Scenario Composition}
\label{subsec:scenario_description}
\noindent \textbf{MDrive-Interdrive.}
This bucket is adopted from the InterDrive benchmark~\cite{colmdriver} without modifications. It contributes 46 interactive multi-agent scenarios, such as intersection crossing, lane merging, and lane changing, in which agents are assigned conflicting or tightly coupled routes, but without any background actors. This bucket serves as an in-domain reference.

\noindent \textbf{MDrive-Interaction.} This bucket contributes 112 scenarios, aiming at evaluating driving systems under three levels of interactivities: 1) \emph{Static Avoidance}, where the driving system needs to traverse through construction zones areas and avoid static obstacles or perform local replan; 2) \emph{Dynamic Avoidance}, where the driving system needs to react to dynamic safety-critical scenarios such as yielding to vulnerable road users (VRUs) in intersection areas; 3) \emph{Dynamic Coordination}, where agent-wise coordination and negotiation is needed for safe and efficient traversal, such as highway merging. The three levels of interactivities are distributed across 11 interaction categories, such as work zones, roundabouts, pedestrian interactions, zipper merges, and overtaking. Each scenario is diverse in terms of limited visibility, occlusion, and collision risk in configurations where single-agent sensing is provably insufficient and scenarios are grounded in NHTSA pre-crash typologies~\cite{najm2007pre}. For each interaction category, scenarios are created by domain experts and are paired with expert driver demonstrations via our human-in-the-loop interface (see Section~\ref{subsec:human_interface}). For example, a human driver completes each scenario via joystick takeover, and the resulting state-control trajectory is logged as a collision-free reference rollout. We then use those human demonstrations to scale our scenarios via an agentic scenario generation pipeline (see Section~\ref{subsec:agentic_gen}), which we leverage an LLM agent to emit a structured schema for scenario creations and convert them as CARLA routes and assets. 

\noindent \textbf{MDrive-V2XPnP.} This bucket is obtained by converting 67 annotated real-world driving logs from the V2XPnP dataset~\cite{zhou2024v2xpnp}, which are characterized by density-heavy intersections and vulnerable-road-user behaviors with multiple CAVs and intelligent infrastructures, into CARLA-executable scenarios via our Real2Sim pipeline (see Section~\ref{subsec:real2sim_pipeline}). The CAV and actor trajectories are reconstructed through a unified scenario description and coordinate transformation, then snapped to the digital-twin lane graph in CARLA for closed-loop execution. This bucket injects multi-agent real-world scenarios and behaviors that are different from previous benchmarks' hand-crafted or simulation-based testing.

Taken together, as \cref{fig:mdrive_representative_scenarios} shows, the three scenario buckets cover coordination-heavy interactive scenarios, perception-limited safety-critical scenarios, balanced language-grounded category expansion, and Real2Sim scenarios. This design allows MDrive to combine targeted preset cases, broader category coverage, and more realistic multi-actor behavior within a single benchmark. 

\begin{wraptable}{r}{0.5\textwidth}
\centering
\footnotesize
\setlength{\tabcolsep}{1.2pt}
\caption{\textbf{MDrive Scenario Distribution}. RL: route length; HC: heading change. Mean actors excludes CAVs and infrastructures.}
\vspace{0.1cm}
\renewcommand{\arraystretch}{1.15} 
\begin{tabular}{lccccc}
\toprule
\textbf{Bucket} & \textbf{Count} & \textbf{CAV} & \textbf{RL~(m)} & \textbf{HC~(deg)} & \textbf{Actors} \\
\midrule
MDrive-InterDrive       & 46  & 3.80 & 51.83 & 65.85 & 0.00 \\
% Pre-crash        & 12  & 2.17 & 43.7 & 56.6 & 1.25 \\
MDrive-Interaction  & 112 & 4.11 & 91.3 & 55.4 & 1.58 \\
MDrive-V2XPnP & 67  & 1.80 & 45.68 & 60.92 & 83.88 \\
\bottomrule
\end{tabular}
% \vspace{0.4cm}
\label{tab:scenario_distribution}
\end{wraptable}

\vspace{-0.1cm}
\subsection{Scenario Statistics and Coverage}

\Cref{tab:scenario_distribution} summarizes the composition of MDrive. The three scenario buckets exhibit distinct structural profiles, and we report the mean number of CAVs, mean CAV route length, mean cumulative heading change, and mean number of background actors. InterDrive has shorter routes and no background actors, while MDrive-Interaction contains longer routes and more actors. MDrive-V2XPnP is distinguished by a higher background actor density than the others, reflecting the real-world traffic.

\begin{figure}[t]
    \centering
    \includegraphics[width=\linewidth]{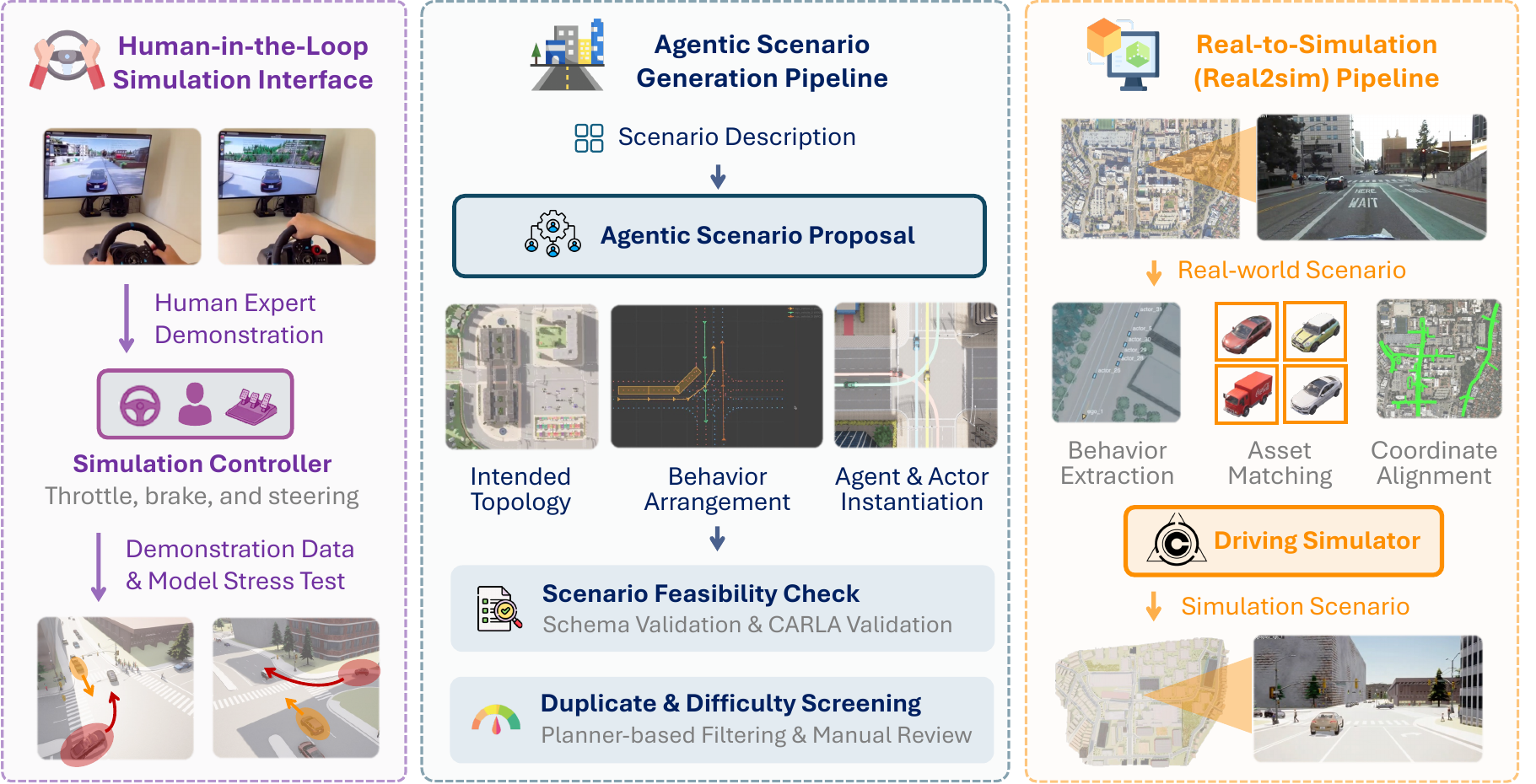}
    \vspace{0.02cm}
    \caption{\textbf{Overview of the MDrive Toolbox.} An open-source toolbox designed to support benchmark construction and extension through three core modules:\textcolor{mdrivepurple}{\textit{1) Human-in-the-Loop Simulation Interface}} that enables expert takeover via physical controllers to collect the realistic demonstrations; \textcolor{mdriveblue}{\textit{2) Agentic Scenario Generation Pipeline}} that leverages agentic system as structured proposers to scalably generate, validate, and curate synthetic multi-agent interactions; and \textcolor{mdriveorange}{\textit{3) Real2Sim Pipeline}} that reconstructs real-world V2X data into closed-loop CARLA scenarios with the digital twin.}
    \label{fig:mdrive_toolbox}
    \vspace{-0.2cm}
\end{figure}
% \vspace{-0.1cm}
\subsection{MDrive Toolbox}
As shown in Fig.~\ref{fig:mdrive_toolbox}, we provide a set of tools to support further explorations with MDrive benchmark and extend the benchmark for customized usage. 

\noindent \textbf{Human-in-the-Loop Simulation Interface}
\label{subsec:human_interface}
This interface allows a human expert to control a CAV in a running scenario while the remaining CAVs and background actors continue to execute their assigned action policies. Thus, the user can create customized scenarios with human demonstrations to perform stress tests for models. We leverage this tool to create initialized scenarios per interaction category as described in \cref{subsec:scenario_description}. The demonstration inputs are provided through a gaming steering wheel with pedals, or a gamepad fallback, mapped to the same throttle, brake, and steering channels that planners use, so that human-driven and policy-driven rollouts are directly comparable, as shown in \cref{fig:mdrive_toolbox}. See Appendix~\ref{subappendix:human_interface} for additional details on the human-in-the-loop interface.

\noindent \textbf{Agentic Scenario Generation Pipeline}
\label{subsec:agentic_gen}
This pipeline is developed to further scale initialized hand-crafted scenarios, as shown in \cref{fig:mdrive_toolbox}. We first give a \emph{scenario description}~\cite{li2023scenarionet, zhao2026bridgesim} template, which includes intended topology, behavior arrangement, and instantiation of relevant actors, by harnessing human instructions and in-context human demonstration data. The language agent proposes several candidate scenarios according to the category descriptions, which are then further validated using human-specified rules in the simulator. The final outputs are exported as scenario directories following CARLA-defined formats. See Appendix~\ref{subappendix:agentic_pipeine} for more details for workflows and orchestrations.

\noindent \textbf{Real2Sim Pipeline}
\label{subsec:real2sim_pipeline}
The MDrive-V2XPnP scenarios are produced by a Real2Sim pipeline that converts the V2XPnP real-world data \cite{zhou2024v2xpnp} into reactive CARLA scenarios. The conversion pipeline consists of a unified scenario description that defines the data format and a digital-twin reconstruction pipeline that aligns real-world and simulator coordinates, creates scenario layouts, and places 3D assets for realistic scenario reconstruction. See Appendix~\ref{subappendix:real2sim_pipeine} for more details of the pipeline.

\subsection{Evaluation Settings}
% MDrive support multi-agent closed-loop driving evaluation with the CARLA simulator. 
We define an MDrive \textit{scenario} as a single executable closed-loop driving episode with fixed map context, CAV assignments, route definitions, dynamic actors, and scene objects. In each scenario, the agent is required to navigate to the target position, taking raw sensor data as input and outputting planning trajectories. In the simulation, both CAV and background actors are reactive agents rolled out in closed-loop manners. Moreover, MDrive also supports open-loop evaluation across perception and planning, and it is simple to customize a new evaluation protocol with our toolkit.

\textbf{Evaluation Metrics.} We adopt the evaluation protocols of InterDrive \cite{winter2025bevdriver} and the CARLA Leaderboard \cite{carla_leaderboard}, using two metrics for closed-loop driving: \textit{(i) Driving Score (DS)}, which measures overall progress and safety based on route completion and infraction scores; and \textit{(ii) Success Rate (SR)}, the percentage of tasks achieving a full driving score. In open-loop evaluation, we refer to the existing multi-agent open-loop evaluation protocols \cite{lei2025risk,lei2025cooperrisk} to adopt Average Precision (AP) with a union threshold of 0.5 to measure perception accuracy, and leverage Average Displacement Error (ADE) and Collision Rate (CR) to measure the planning safety and accuracy. See~\cref{appendix:mdrive_detailed_settings} for more details.

\subsection{Scope and Sim2Real Gap Interpretation}
\label{subsec:scope_sim2real}
\textbf{MDrive Scope.} Closed-loop cooperative driving raises a class of scientific questions, \textit{i.e.}, how perception sharing, decision negotiation, and reactive background traffic jointly shape multi-agent planning, that can only be examined under reproducible and safety-controllable conditions. MDrive is built as a principled testbed for this purpose. 
Within a CARLA-based environment, MDrive enables reactive multi-agent rollouts, controlled ablations across cooperation paradigms, and broad coverage of long-tail interactions grounded in NHTSA pre-crash typologies~\cite{najm2007pre}, all of which are difficult to obtain at scale and with safety guarantees in on-road data collection. 
To keep the behavioral side of the loop realistic, our Real2Sim pipeline (Section~\ref{subsec:real2sim_pipeline}) ports real-world V2XPnP driving logs~\cite{zhou2024v2xpnp} into reactive CARLA scenarios, so that background traffic, vulnerable road users, and intersection topologies inherit the distributions of real urban driving rather than hand-crafted heuristics. 
MDrive-Toolbox is intentional modular, so that future improvements in sensor models, V2X channel modeling, and scene rendering can be customized based on the community's research scopes, without redesigning the benchmark.

\textbf{Sim2Real Gap.} Though building MDrive on CARLA inevitably introduces a Sim2Real gap, the principal value of MDrive lies in its ability to isolate causal questions about cooperation.
MDrive evaluates every policy on identical maps, routes, traffic, and weather across three complementary buckets. 
It allows each performance change to be attributed to a single targeted intervention, \textit{e.g.}, (\emph{i}) single-agent versus multi-agent operation; (\emph{ii}) perception sharing versus decision negotiation; or (\emph{iii}) open-loop versus closed-loop evaluation, while holding all other factors fixed.
While we do not argue that results from MDrive transfer directly to real-world deployment, we position MDrive as a valuable hypothesis testing platform and enables controlled analysis for further methodological developments from the community. The resulting insights, \textit{i.e.}, that closed-loop training distributions matter, that language-grounded negotiation must remain context-aware in dense traffic, and that open-loop metrics underdetermine closed-loop behavior (Sections~\ref{sec:benchmarking}--\ref{subsec:ol_cl}), therefore describe properties of cooperative driving systems themselves rather than artifacts of any particular simulator.
These conclusions are intended to guide the design and evaluation of multi-agent driving systems. 
We position MDrive as a living foundation that can absorb future advances in sensor realism, V2X communication modeling, and rendering fidelity as the Sim2Real distance continues to narrow.

% \rowcolor[HTML]{fff3e9}
\section{Benchmarking State-of-the-Art Models}
\label{sec:benchmarking}
In this section, we present benchmark results to answer two research questions (RQ):
\begin{itemize}[leftmargin=*]
    \item \textbf{RQ1:} \textit{To what extent does multi-agent cooperation improve closed-loop driving over single-agent systems, and how do the gains decompose between perception sharing and negotiation?}
    \item \textbf{RQ2:}\textit{ What performance gaps and new challenges for multi-agent system do MDrive expose?}
\end{itemize}

\vspace{-0.1cm}
We benchmark on a set of representative single-agent and multi-agent end-to-end driving models. Single-agent driving models include UniAD~\cite{uniad}, VAD~\cite{vad}, TCP~\cite{wu2022trajectory}, LMDrive~\cite{shao2024lmdrive}, multi-agent driving models include CoDriving~\cite{codriving}, CoLMDriver~\cite{colmdriver} and its rule-based variant.

\begin{figure}[b]
    \centering
    \vspace{-0.4cm}
    \includegraphics[width=.95\linewidth]{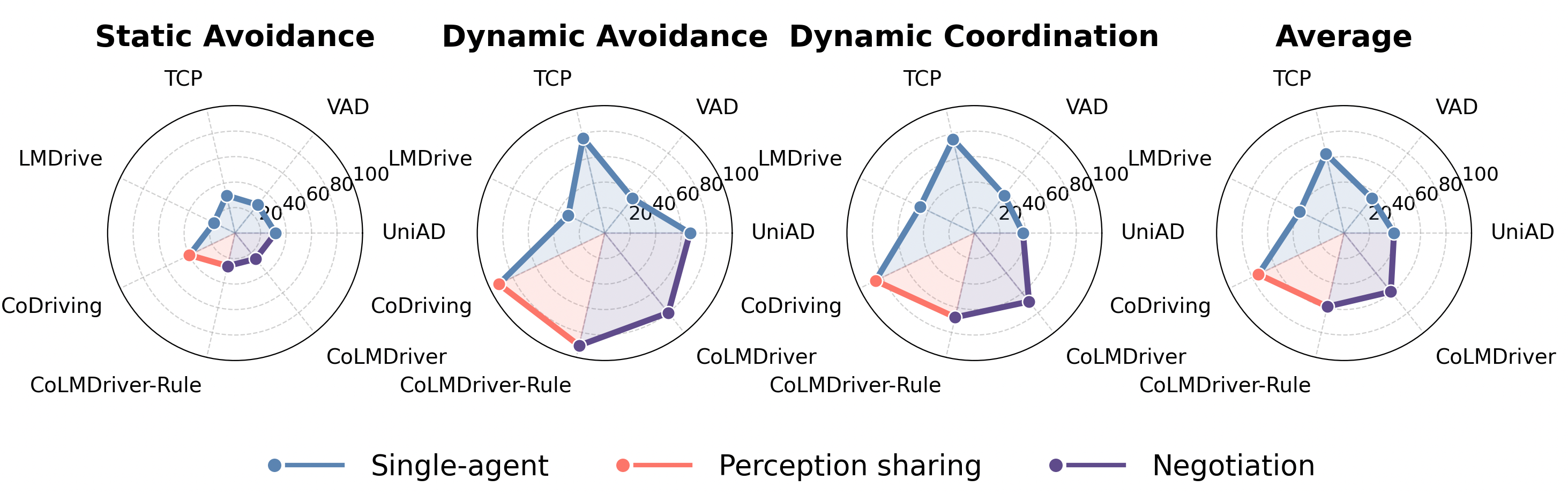}
    \caption{\textbf{Category-wise MDrive Benchmark.} Harmonic mean between Driving Score and Success Rate on three high-level categories. }
    \label{fig:mdrive_highlevel_subscore}
    \vspace{-0.4cm}
\end{figure}

\subsection{RQ1: Where Cooperation Helps}
\label{subsec:rq1}
We answer RQ1 by first measuring whether cooperation benefits the final planning task in aggregate, then decomposing the gain along the two dimensions: perception sharing and decision negotiation. We reserve causal diagnosis of where cooperation \emph{fails} for \Cref{subsec:rq2}.

\noindent \textbf{Cooperation enhances closed-loop driving non-trivially.} According to \cref{tab:mdrive_results}, multi-agent models generally perform better than single-agent systems in terms of DS and SR across different subsets. \cref{fig:mdrive_highlevel_subscore} shows the multi-agent$-$single-agent gap rises from 3.54 on \textit{Static Avoidance} scenarios to 35.12 on \textit{Dynamic Avoidance} scenarios and 24.58 on \textit{Dynamic Coordination} scenarios in terms of Harmonic mean between DS and SR. The gain concentrates exactly where the ego must reason about other agents' intent under partial observability.

\noindent \textbf{Perception sharing prevents planning failure from lack of observability.} CoDriving posts the highest harmonic mean on all three interactivity levels, as \cref{fig:mdrive_highlevel_subscore} illustrates, leading the next-best multi-agent system by 1.68 on \textit{Dynamic Avoidance} scenarios and 10.63 on \textit{Dynamic Coordination} scenarios. Cooperative perception expands the observable region beyond ego sensors, which leads to better perceptual abilities that prevent policies failing due to lack of observability.

\noindent \textbf{Negotiation benefits dynamic coordination.} As\cref{tab:mdrive_results} shows, CoLMDriver beats all the baselines on the negotiation-specific subset MDrive-InterDrive without background actors, and beat most of the single agent baselines on \textit{Dynamic Coordination} scenarios by 27.7, but trail CoDriving. That indicates that negotiation works best when the conflict pair is evident and exclusive among CAVs only, but degradation occurs once the environment becomes complicated. We trace the cause to a benchmark-revealed failure mode in \Cref{subsec:rq2}.

\begin{table}[t]
\centering
\caption{\textbf{MDrive Benchmark Across Three Scenario Buckets.} Avg: mean across all the scenarios. *: rule-based planner variant. \textbf{Bold}, \underline{underline}, and \uwave{wavy} marks first, second, third place, respectively.}
\vspace{0.2cm}
\footnotesize
\label{tab:mdrive_results}
\renewcommand{\arraystretch}{1.2}
\setlength{\tabcolsep}{6.5pt}
\definecolor{customColor}{HTML}{FFF3E9} 
\begin{tabular}{l|cc|cccccc|cc}
\toprule
\multirow{3}{*}{\textbf{Method}} 
 & \multirow{3}{*}{\rotatebox{90}{\footnotesize \textbf{Perc.Sharing}}}
 & \multirow{3}{*}{\rotatebox{90}{\footnotesize \textbf{Negotiation}}}
 & \multicolumn{8}{c}{\raisebox{-.18\height}{\includegraphics[width=0.07\textwidth]{figure/MDrive-Logo.png}} \hspace{0.0001em}\textbf{MDrive}} \\
\cmidrule(lr){4-11}
 & & & \multicolumn{2}{c}{MDrive-InterDrive} 
       & \multicolumn{2}{c}{MDrive-Interaction} 
       & \multicolumn{2}{c}{MDrive-V2XPnP} 
       & \multicolumn{2}{|c}{\textbf{Avg}}  \\
\cmidrule(lr){4-5} \cmidrule(lr){6-7} \cmidrule(lr){8-9} \cmidrule(lr){10-11}
 & &
 & DS↑ & SR↑
 & DS↑ & SR↑
 & DS↑ & SR↑
 & DS↑ & SR↑ \\
\midrule
\rowcolor[HTML]{eaf1f8} \multicolumn{11}{l}{\textit{Single-agent Models}} \\
TCP~\cite{wu2022trajectory}  & $\times$ & $\times$ & 83.55 & 67.80 & \underline{75.80}& \underline{55.25}& 71.02 & 58.70 & \uwave{76.79} & 60.58\\
UniAD~\cite{uniad}  & $\times$ & $\times$ & 57.48 & 26.40 & 53.08& 31.10& 72.55 & 68.48 & 61.04& 41.99\\
VAD~\cite{vad}  & $\times$ & $\times$ & 57.49 & 23.30 & 48.83& 27.14& 67.20 & 66.30 & 57.84& 38.91\\
LMDrive~\cite{shao2024lmdrive}  & $\times$ & $\times$ & 56.74 & 27.10 & 57.45& 29.08& 71.56& 69.57& 61.92& 41.92\\
\midrule
\rowcolor[HTML]{fff3e9} \multicolumn{11}{l}{\textit{Multi-agent Models}} \\
CoDriving~\cite{codriving}  & \checkmark & $\times$   & \underline{83.99} & 66.10 & \textbf{79.06}& \textbf{70.53}& \textbf{84.97} & \textbf{82.84}& \textbf{82.67}& \textbf{73.16}\\
CoLMDriver*~\cite{colmdriver}  & \checkmark & \checkmark & 83.77 & \underline{75.10} & 66.70& 53.65 & 77.89& 73.13& 76.12& \uwave{67.29}\\
CoLMDriver~\cite{colmdriver}  & \checkmark & \checkmark & \textbf{85.29} & \textbf{76.85} & 67.91 & 51.93 & \underline{78.27} & \underline{75.37}& \underline{77.16} & \underline{68.05}\\
\bottomrule
\end{tabular}
% \vspace{-0.4cm}
\end{table}

\subsection{RQ2: V2X Performance Gaps Revealed by MDrive}
\label{subsec:rq2}
According to \cref{tab:mdrive_results}, perception-sharing-based systems such as CoDriving~\cite{codriving} and negotiation-based systems such as CoLMDriver~\cite{colmdriver} generally perform better than their single-agent counterparts, but discrepancies also occur where multi-agent systems might also fail to leverage the provided information from other agents. In this section, we further analyze the potential limitations of current multi-agent systems and shed light on future work that improves the current bottlenecks. 

\begin{figure}[tbh]
    \centering
    \includegraphics[width=\linewidth]{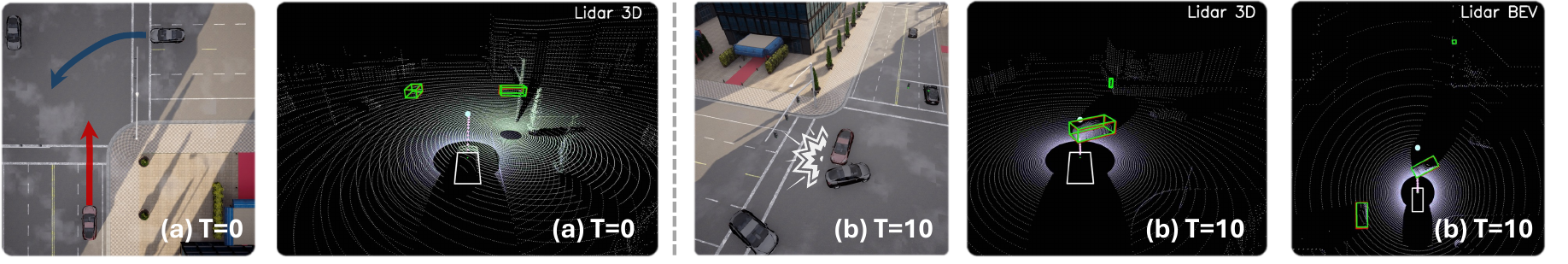}
    \caption{\textbf{Failure Mode: Perception Sharing does not Guarantee Better Planning.} The ego CAV receives the detections of the collision object at T=0, but still cannot avoid it at T=10.}
    \label{fig:codriving_failure}
    % \vspace{-0.2cm}
\end{figure}

\noindent \textbf{Perception sharing doesn't guarantee improved planning.} Perception sharing is the current mainstream in V2X cooperative driving, and most methods follow the assumption that better perception might implicitly translate into better planning, although we later show in~\Cref{subsec:ol_cl} that this \emph{open-loop perception to closed-loop planning correlation does not always hold}. Across \cref{tab:mdrive_results} and \cref{fig:mdrive_highlevel_subscore}, CoDriving consistently outperforms all single-agent baselines, with the largest gains in dynamic categories where shared perception of CAVs reduces ego planning uncertainties by providing more observability. However, this does not necessarily enable the model to mitigate covariate shift or exhibit corrective behaviors. As illustrated in Fig.~\ref{fig:codriving_failure}, ego planning can fail due to a mismatch between state visitations and the training distribution, despite detecting the collision object at the beginning. For future works to bridge this gap, it's necessary to train the driving policy with a closed-loop paradigm and extend the V2X perception benefits to the final closed-loop driving. 

\noindent \textbf{Negotiation difficulty increases substantially in complex and dense traffic scenarios.} Negotiation offers an alternative cooperative path, where agents resolve conflicts through explicit communication rather than shared perception. Across \cref{tab:mdrive_results}, CoLMDriver and its rule-based variant beat all single-agent baselines on the negotiation-rich subset (i.e., MDrive-InterDrive) but never match CoDriving's average. The degradation is more significant when we introduce rich background actors. We attribute this to the fact that the LLM negotiation process of CoLMDriver is easily distracted by complicated background interactions, which are not revealed, since the InterDrive benchmark doesn't have those background actors. In these cases, there might not be effective agent-wise negotiations, harming the driving safety and efficiency. We further conduct a human study in \Cref{subsec:human_study} to characterize where the gap actually lies and what behaviors current LLM negotiators miss.

\section{Discussion}
Building on the results in Section~\ref{sec:benchmarking}, we conduct additional experiments on multi-agent driving systems around the open-loop-to closed-loop (OL-CL) evaluation gap and evaluate the quality of the specific cooperative behavior with negotiation based on human demonstration:
\begin{itemize}[leftmargin=*]
    \item \textbf{RQ3:} \textit{How large is the gap between open-loop and closed-loop evaluation in multi-agent systems?}
    \item \textbf{RQ4:} \textit{What is the quality of cooperative behavior for existing driving models with negotiation?}
\end{itemize}

\begin{figure}[tbh]
    \centering
    \includegraphics[width=\linewidth]{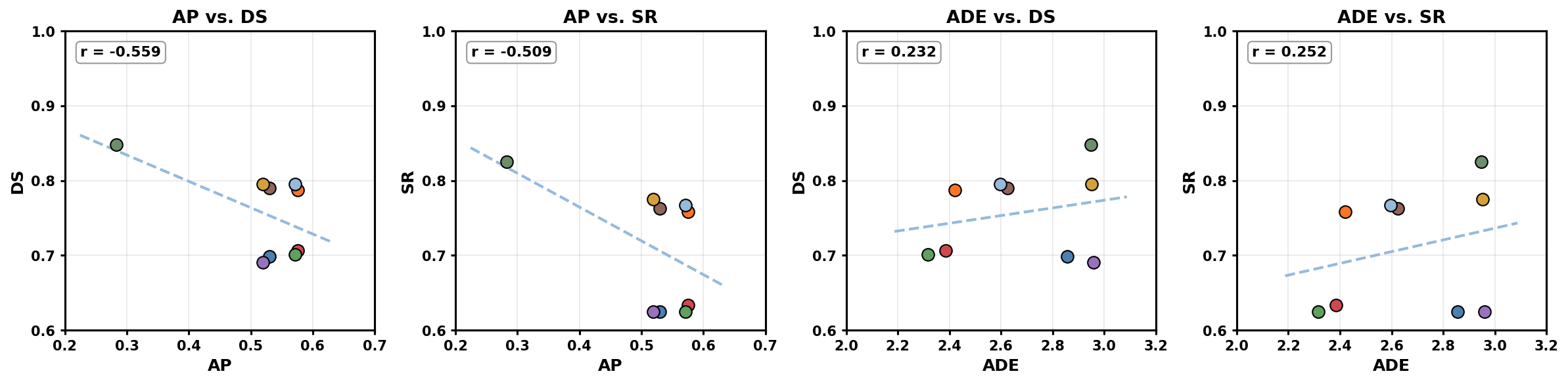}
    \caption{\textbf{Correlation of Open-loop to Closed-loop Evaluation.} We compare the open-loop metrics, such as AP50 and ADE, against closed-loop metrics, such as DS and SR, for a set of 9 planners. }
    \vspace{-0.2cm}
    \label{fig:mdrive_olcl_correlation}
    % \vspace{-0.2cm}
\end{figure}

\subsection{RQ3: Open-loop to Closed-loop Evaluation Gap}
\label{subsec:ol_cl}

The experiments in this section are designed to highlight the necessity of closed-loop evaluation. Open-loop evaluations directly leverage static real-world driving logs and are straightforward and cheap to implement~\cite{caesar2020nuscenes, caesar2021nuplan, dauner2024navsim, li2024pretrain}, but driving is inherently a closed-loop task within interactive environments. We analyze the correlation between open-loop and closed-loop metrics to demonstrate the evaluation gap. For open-loop evaluation, we adopt widely used metrics, including detection AP and planning trajectory L2 error~\citep{zhou2024v2xpnp, lei2025risk, zhao2025quantv2x, zhao2024coopre}. For closed-loop evaluation, we report DS and SR. Figure~\ref{fig:mdrive_olcl_correlation} plots open-loop metrics against closed-loop metrics on MDrive-V2XPnP. We find that for multi-agent systems, stronger perception or prediction performance does not account for stronger closed-loop performance. This observation indicates that open-loop metrics can only serve as a rough proxy for the closed-loop driving task, which is consistent with recent studies on the evaluation gap in single-agent system~\cite{zhao2026bridgesim, ol_cl_survey}. Therefore, closed-loop evaluation is essential for benchmarking multi-agent driving systems and avoiding misleading conclusions about planning performance. See~\Cref{subappendix:ol_cl_exp} for more experimental details and outlier analysis.

\begin{figure}[tbh]
    \centering
    % \vspace{-0.2cm}    
    \includegraphics[width=\linewidth]{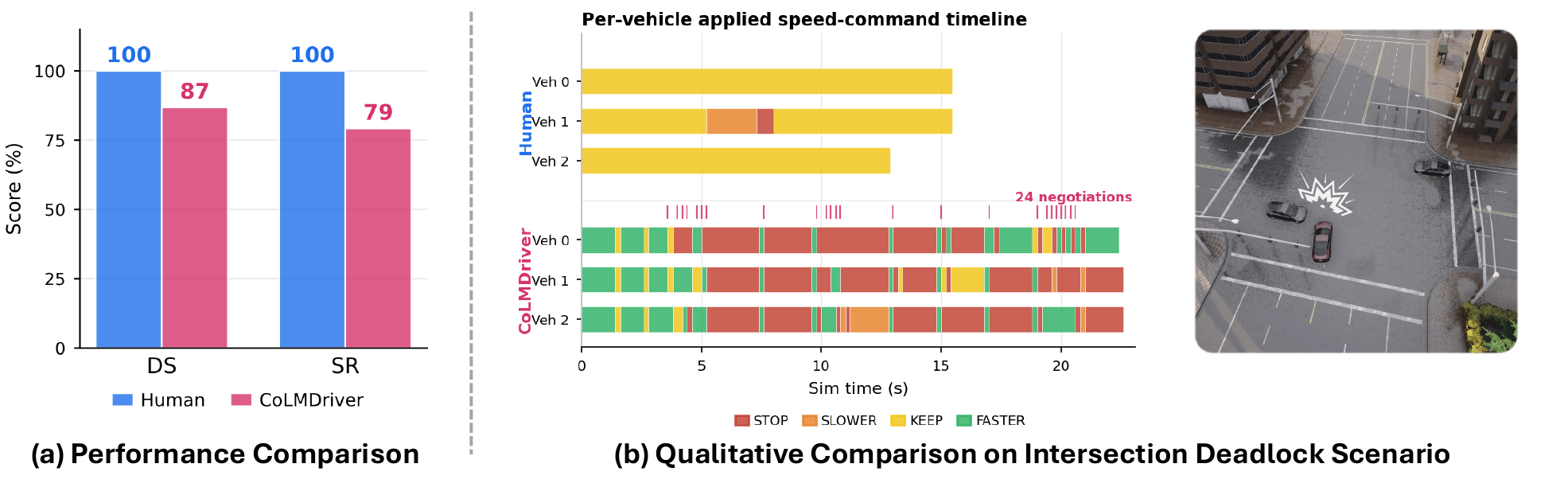}
    \caption{\textbf{Cooperative Behavior Quality Evaluation with Human Demonstration.} \textbf{(a)}: statistics of metrics comparison between CoLMDriver and Human; \textbf{(b)}: Visualization of where CoLMDriver fails under complex and ambiguous situations.}
    \label{fig:mdrive_human_experiments}
\end{figure}

\subsection{RQ4: Cooperative Behavior Quality}
\label{subsec:human_study}
We conduct a human study comparing human experts against CoLMDriver on representative scenarios to quantify the quality of the specific cooperative behavior with negotiation using human-likeness in driving metrics. As shown in ~\Cref{fig:mdrive_human_experiments}, we report DS, SR for both human experts and CoLMDriver. The results indicate that the primary gap lies in the success rate, where the multi-agent system frequently fails to reach navigation goals. In addition, CoLMDriver exhibits lower efficiency during multi-agent negotiation.
%, resulting . 
These findings highlight that future work might develop human-like negotiation strategy to improve closed-loop planning performance in context-rich settings. See Appendix~\ref{subappendix:human_study} for more experimental details.

\section{Conclusion}
We propose \textbf{MDrive}, a closed-loop cooperative driving benchmark with diverse interaction scenarios, and systematically analyze the V2X benefits across perception sharing and decision negotiation. MDrive includes 225 scenarios organized into three complementary categories, grounded in both NHTSA pre-crash typologies and real-world V2X datasets. We further provide a toolbox to support human-in-the-loop simulation, agentic scenario generation, and a Real2Sim conversion that ensures broad feasibility and compatibility. MDrive also reveals the bottlenecks in current multi-agent methods, encouraging future work on resolving the closed-loop generalization gap. We hope MDrive provides the community with a reproducible foundation for diagnosing closed-loop generalization in cooperative autonomous driving and for developing the next generation of multi-agent systems.

\paragraph{Limitations and Future Works.} MDrive is built on CARLA and inherits its sensor, dynamics, and rendering fidelity, which still leaves a non-trivial sim-to-real gap relative to on-road deployment. Future work includes: 1) Better integration of cooperative perception and negotiation within a single end-to-end policy, and 2) Closed-loop training and adaptation methods that explicitly target the long-tailed safety-critical interactions where current multi-agent systems still degrade most sharply.

% show some summary and key insights here
\bibliographystyle{unsrtnat}
\bibliography{ref}

%%%%%%%%%%%%%%%%%%%%%%%%%%%%%%%%%%%%%%%%%%%%%%%%%%%%%%%%%%%%
\newpage
\appendix
\setcounter{section}{0}
\setcounter{figure}{0}
\setcounter{table}{0}
\setcounter{equation}{0}
\renewcommand{\thesection}{\Alph{section}}
\renewcommand{\thefigure}{S\arabic{figure}}
\renewcommand{\thetable}{S\arabic{table}}
\renewcommand{\theequation}{S\arabic{equation}}

\begin{center}
\LARGE \textbf{\raisebox{-.18\height}{\includegraphics[width=0.0853\textwidth]{figure/MDrive-Logo.png}} \hspace{0.0001em}%
  \textit{MDrive} Appendices}
\end{center}

\let\oldaddcontentsline\addcontentsline
\let\oldaddtocontents\addtocontents

\renewcommand{\addcontentsline}[3]{}
\renewcommand{\addtocontents}[2]{}

\begingroup
\newcommand{\inst}[1]{}
\author{}
% \institute{}
\maketitle
\endgroup

\let\addcontentsline\oldaddcontentsline
\let\addtocontents\oldaddtocontents
\vspace{0.75cm}
% --- NEW FIX: Isolate Supplementary TOC into a completely separate .stc file ---
\makeatletter
% 1. Redirect all subsequent TOC writing from .toc to .stc
\let\oldaddcontentsline\addcontentsline
\renewcommand{\addcontentsline}[3]{%
    \def\@tempa{#1}\def\@tempb{toc}%
    \ifx\@tempa\@tempb
        \oldaddcontentsline{stc}{#2}{#3}%
    \else
        \oldaddcontentsline{#1}{#2}{#3}%
    \fi
}
\let\oldaddtocontents\addtocontents
\renewcommand{\addtocontents}[2]{%
    \def\@tempa{#1}\def\@tempb{toc}%
    \ifx\@tempa\@tempb
        \oldaddtocontents{stc}{#2}%
    \else
        \oldaddtocontents{#1}{#2}%
    \fi
}

% 2. Keep the bold formatting for Level 1 headings
\let\old@l@section\l@section
\renewcommand{\l@section}[2]{\old@l@section{\textbf{#1}}{\textbf{#2}}}
\makeatother
% -------------------------------------------------------------------------------

{
    \hypersetup{linkcolor=black}
    \begin{spacing}{1} 
        \setcounter{tocdepth}{2} 
        \renewcommand{\contentsname}{} 
        
        \makeatletter
        % 3. Redefine tableofcontents to solely read from our clean .stc file
        \renewcommand\tableofcontents{%
            \@starttoc{stc}%
        }
        \makeatother
        
        \tableofcontents
    \end{spacing}
}

\vspace{0.5cm}

% \section{More Related Works}
% \textbf{End-to-end Autonomous Driving.} 
% Traditional autonomous driving systems follow a modular paradigm with independent perception \cite{li2024bevformer,lang2019pointpillars}, prediction \cite{shi2023mtr,zhou2023qcnext,zhou2022comprehensive}, and planning \cite{zhou2021reliable,huang2024gen} components. While this structure is easier to design and maintain, it is limited by error accumulation across modules \cite{chen2024end}. In contrast, end-to-end approaches have become the mainstream, which eliminate inter-module error propagation, simplify the system, and enable joint optimization for the final planning task \cite{DrivoR,RAP,zhou2025autovla,zhou2026spanvla}. However, current paradigms primarily focus on single-agent systems and remain constrained by local perception and passive planning \cite{diffusiondrivev2,LEAD}. Long-tail scenarios have become a key focus in autonomous driving \cite{gerstenecker2026fail2drive,waymo2025e2e}. However, the potential of multi-agent systems, i.e., perception sharing and decision negotiation, especially in scenarios where existing systems fail, remains underexplored.

\section{Additional Details of MDrive Scenarios}
\subsection{Scenario Description}
\textbf{MDrive-Interaction.} \cref{fig:mdrive_representative_scenarios} shows the typical scenarios for 11 interaction categories, and their specific description are provided as following:
\begin{enumerate}[leftmargin=*]
    \item \textit{Pre-crash}: These scenarios are crafted and grounded by National Highway Traffic Safety Administration (NHTSA) guidance, focusing on challenging scenarios with occlusion, limited perception range, and dangerous interaction behavior.
    \item \textit{Blocked Lane Obstacle}: Diverse static obstacle block the travel lane.
    \item \textit{Construction Zone}: Work-zone lane constriction. 
    \item \textit{Highway On-Ramp Merge}: Ramp merges into highway. 
    \item \textit{Interactive Lane Change}: Multi-lane weaving interactions.
    \item \textit{Intersection Deadlock Resolution}: Uncontrolled multi-vehicle right-of-way.
    \item \textit{Major / Minor Unsignalized Entry}: Minor road enters major at unsignalized junction.
    \item \textit{Overtaking on Two-Lane Road}: Pass the vehicle or obstacles with oncoming traffic.
    \item \textit{Pedestrian Crosswalk}: Pedestrians crossing vehicle travel path.
    \item \textit{Roundabout Navigation}: Multiple CAVs should yield, merge, and travel through the roundabout.
    \item \textit{Unprotected Left Turn}: CAVs need to left turn through oncoming traffic.
\end{enumerate}
For the high level interaction category, the \textit{Dynamic Coordination} includes 1,4,5,6,7,10,11, and the \textit{Static Avoidance} covers 2,3,8, and the rest are the \textit{Dynamic Avoidance}.

For weather and illumination conditions, each category is evenly distributed across 3 cloudy scenarios, 3 nighttime scenarios, 2 default-condition scenarios, and 2 rainy scenarios, as \cref{fig:mdrive_representative_scenarios} shown. These settings can be easily customized by modifying the weather presets in the scenario XML descriptions.

\begin{figure}[h]
    \centering
    \includegraphics[width=.99\linewidth]{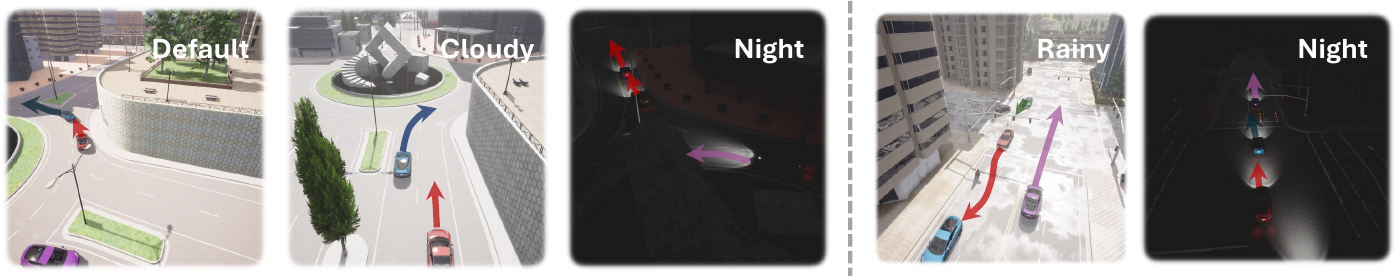}
    \caption{\textbf{Various Weather and Illumination Conditions} for each interaction category of MDrive-Interaction. Left: Default-condition, Cloudy, and Night conditions for the roundabout scenario. Right: Rainy and Night condition for construction zone.}
    \label{fig:various_weather_conditions}
\end{figure}

\begin{figure}[h]
    \centering
    \includegraphics[width=.99\linewidth]{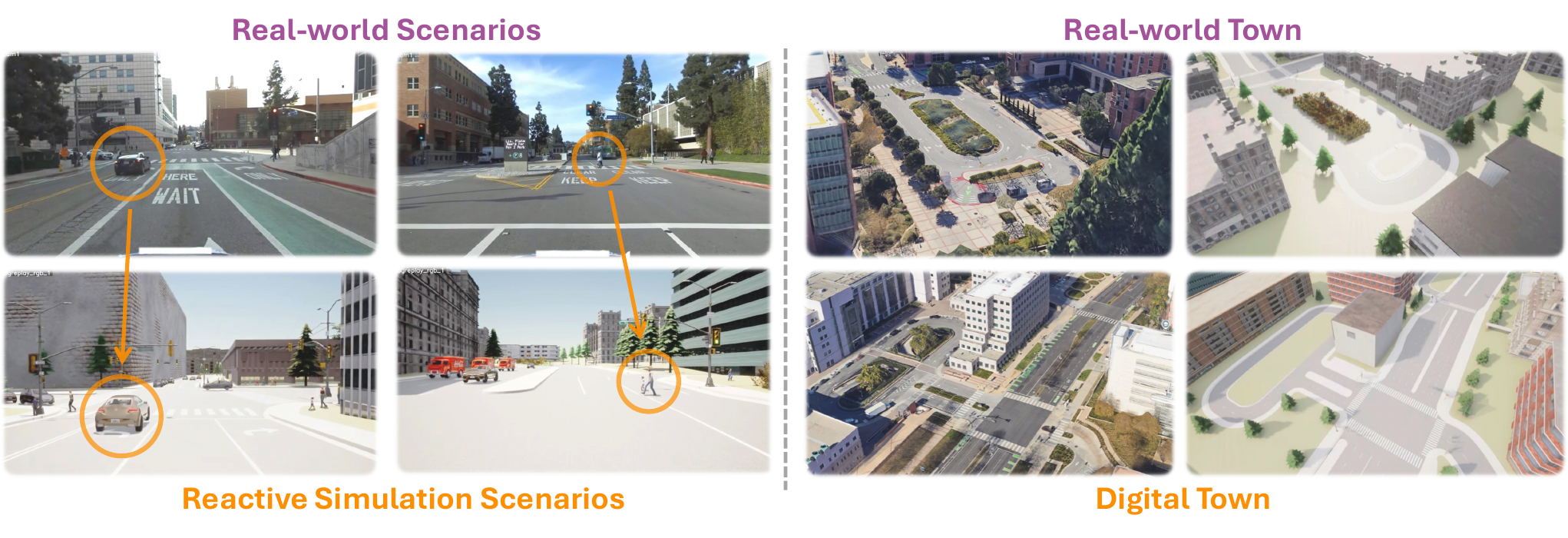}
    \caption{\textbf{Real2Sim Visualization}. Left: two Real2Sim scenarios demos. Right: Real2Sim Town.}
    \label{fig:real2sim_demo}
\end{figure}

\textbf{MDrive-V2XPnP.} To convert the V2XPnP real-world scenarios \cite{zhou2024v2xpnp} to reactive CARLA scenarios, we follow the pipeline in \cref{subappendix:real2sim_pipeine} to build the digital town for the dataset. \cref{fig:real2sim_demo} shows some Real2Sim scenario and digital town comparison.

\subsection{Additional Scenario Statistics}
\cref{tab:scenario_distribution} show the data distribution across three data buckets, i.e., MDrive-InterDrive, MDrive-Interaction, and MDrive-V2XPnP. \cref{tab:llm_category_breakdown} further breaks down the MDrive-Interaction subset by interaction category. While the category counts are uniform, the structural statistics vary across categories. For example, roundabout navigation has the largest mean cumulative heading change, unprotected left turn contains the highest mean number of CAV vehicles, and construction-zone scenarios include more background actors on average than several other categories. 
% These differences show that the balanced MDrive-Interaction subset is not structurally uniform across categories.

\begin{table*}[t]
\centering
\footnotesize
\setlength{\tabcolsep}{7pt}
\renewcommand{\arraystretch}{1.1}
\caption{\textbf{MDrive-Interaction Subset Distribution.} The statistic is broken down by scenario category. Mean actors excludes CAVs.}
\begin{tabular}{l r r r r r}
\toprule
\makecell[l]{\textbf{Interaction Category}} & \textbf{Count} & \makecell[c]{\textbf{CAV}} & \makecell[c]{\textbf{Route Length (m)}} & \makecell[c]{\textbf{Heading Change (deg)}} & \textbf{Actors} \\
\midrule
Pre-crash & 12 & 2.17 & 42.6  & 91.5  & 1.25 \\
Blocked Lane Obstacle & 10 & 4.70 & 88.8  & 6.5   & 3.40 \\
Construction Zone & 10 & 4.60 & 95.5  & 6.4   & 7.10 \\
Highway On-Ramp Merge & 10 & 4.60 & 121.3 & 18.5  & 0.00 \\
Interactive Lane Change & 10 & 4.20 & 100.9 & 43.5  & 0.00 \\
Intersection Deadlock Resolution & 10 & 4.70 & 94.5  & 89.7  & 0.00 \\
Major / Minor Unsignalized Entry & 10 & 4.00 & 71.4  & 34.0  & 0.60 \\
Overtaking on Two-Lane Road & 10 & 3.70 & 63.0  & 0.2   & 2.10 \\
Pedestrian Crosswalk & 10 & 3.80 & 92.5  & 9.5   & 2.70 \\
Roundabout Navigation & 10 & 4.00 & 112.1 & 296.3 & 0.00 \\
Unprotected Left Turn & 10 & 5.10 & 95.9  & 40.2  & 0.30 \\
\bottomrule
\end{tabular}
\label{tab:llm_category_breakdown}
\end{table*}

\section{MDrive Closed-loop Simulation Details}
\label{appendix:mdrive_detailed_settings}
\subsection{Simulation Setting}
\label{appendix:simulation_set}
The MDrive is implemented on the CARLA simulator of version 0.9.12 \cite{carla} in NVIDIA L40S GPUs. Our evaluation baselines spanning three agent classes: cooperative models (i.e., CoLMDriver, CoLMDriver-rulebase, and CoDriving), perception-swap models (i.e., rule-based planner paired with cooperative detectors AttFuse \cite{xu2022opv2v}, DiscoNet \cite{disconet}, F-Cooper \cite{chen2019f}, and CoBEVT \cite{xu2022cobevt}),  single-vehicle end-to-end models (i.e., TCP, UniAD, VAD, LMDrive).

\subsection{Sensor Setting}
% Our evaluation uses ten baselines spanning four agent classes:
% cooperative V2V (CoLMDriver, CoLMDriver-rulebase, CoDriving),
% perception-swap (with the basic-agent planner, paired with detectors
% AttFuse, DiscoNet, F-Cooper, and CoBEVT), the upstream
% OpenCDA baseline (perception-swap detectors paired with OpenCDA's
% BehaviorAgent and PID), human-in-the-loop, and four single-vehicle
% end-to-end baselines (TCP, UniAD, VAD, LMDrive).  
All agents share a co-located GNSS\,/\,IMU pair at the vehicle origin and a \(20\,\text{Hz}\) speedometer pseudo-sensor.  LiDAR ray-cast attributes are fixed by the leaderboard wrapper, with a default profile (64 channels, \(85\,\text{m}\) range, \(20\,\text{Hz}\), \(5\!\times\!10^{5}\) pts/s, vertical FOV \([-30^{\circ},+10^{\circ}]\)) and a special-case profile for LMDrive (\(10\,\text{Hz}\), \(6\!\times\!10^{5}\) pts/s; same range, channels, and vertical FOV). The wrapper overrides any agent-declared ray attributes; only the LiDAR pose is honoured.

\begin{table*}[h]
\centering
\small
\caption{Per-agent sensor configuration.  Camera entries list count,
resolution, and FOV; the four-camera rigs use yaws
\(\{0^{\circ},180^{\circ},-60^{\circ},+60^{\circ}\}\) at
\(x=\pm 1.3\,\text{m}\), \(z=2.3\,\text{m}\) unless noted.  Multi-camera
nuScenes-style rigs (UniAD, VAD) use \(70^{\circ}\) FOV for front and
side cameras, \(110^{\circ}\) FOV for the back camera, mounted at
\(z=1.6\,\text{m}\).}
\setlength{\tabcolsep}{15pt}
\renewcommand{\arraystretch}{1.1} 
\label{tab:sensors}
\begin{tabular}{lll}
\toprule
\textbf{Agent (baselines)} & \textbf{RGB cameras} & \textbf{LiDAR} \\
\midrule
\texttt{colmdriver\_agent}
  & 4\,$\times$\,\(3000\!\times\!1500\), \(100^{\circ}\)
  & \checkmark \\
\quad CoLMDriver & & \\[2pt]
\texttt{pnp\_agent\_e2e\_v2v}
  & 4\,$\times$\,\(1600\!\times\!900\), \(100^{\circ}\)
  & \checkmark \\
\quad CoDriving & & \\[2pt]
\texttt{perception\_swap\_agent}
  & front \(800\!\times\!600\)/\(100^{\circ}\) 
  & \checkmark \\
\quad Perception-swap 
  & top-BEV \(800\!\times\!800\)/\(90^{\circ}\) , pitch \(-90^{\circ}\)
   \\[2pt]
\texttt{hitl\_agent}
  & 4\,$\times$\,\(640\!\times\!360\), \(100^{\circ}\)
  & --- \\
\quad human study & & \\[2pt]
\texttt{tcp\_agent}
  & front \(900\!\times\!256\)/\(100^{\circ}\) 
  & --- \\
\quad TCP
  & top-BEV \(256\!\times\!256\)/\(120^{\circ}\) , pitch \(-90^{\circ}\)
  & \\[2pt]
\texttt{uniad\_b2d\_agent}
  & 6\,$\times$\,\(1600\!\times\!900\) (nuScenes rig)
  & --- \\
\quad UniAD
  & top-BEV \(512\!\times\!512\)/\(50^{\circ}\) , pitch \(-90^{\circ}\)
  & \\[2pt]
\texttt{vad\_b2d\_agent}
  & 6\,$\times$\,\(800\!\times\!450\) (nuScenes rig)
  & --- \\
\quad VAD & & \\[2pt]
\texttt{lmdriver\_agent}
  & front \(1200\!\times\!900\)/\(100^{\circ}\) 
  & \checkmark\\
\quad LMDrive
  & rear/left/right \(400\!\times\!300\)/\(100^{\circ}\), same \(z\)
   \\
\bottomrule
\end{tabular}
\\[2pt]
\end{table*}

\subsection{Controller Setting}
Each baseline uses its own low-level controller, and we do not impose a uniform tracker.  All controllers are invoked at the CARLA simulation rate of \(20\,\text{Hz}\) (\(\Delta t=0.05\,\text{s}\) per tick). Throttle outputs are continuous in \([0,\,0.75]\); brake conventions vary (see column).  Steer outputs are continuous, with the listed clip.

\begin{table*}[h]
\centering
\scriptsize
\caption{Per-baseline controller configuration.  Gains are
\((K_P,K_I,K_D)\); ``adaptive'' target speeds are computed online from
the planner's predicted waypoints and capped by \texttt{max\_speed}.
``binary brake'' = brake \(\in\{0,1\}\) triggered by speed-vs-target
threshold; ``cont.'' = continuous brake in \([0,1]\).}
\label{tab:controllers}
\setlength{\tabcolsep}{0.7pt}
\renewcommand{\arraystretch}{1.4} 
\begin{tabular}{lcccccc}
\toprule
\textbf{Baseline} & \textbf{Controller} & \textbf{Long. gains} & \textbf{Lat. gains}
         & \textbf{Throttle / Brake max} & \textbf{Steer clip} & \textbf{Target speed} \\
\midrule
CoLMDriver
  & \texttt{V2X\_Controller}
  & \((5.0,\,1.0,\,0.1)\) & \((1.0,\,0.2,\,0.1)\)
  & \(0.75\) / binary  & \(\pm1.0\)
  & adaptive, cap \(8\,\text{m/s}\) \\
CoLMDriver*
  & \texttt{V2X\_Controller}
  & \((5.0,\,1.0,\,0.1)\) & \((1.0,\,0.2,\,0.1)\)
  & \(0.75\) / binary  & \(\pm1.0\)
  & adaptive, cap \(8\,\text{m/s}\) \\
CoDriving
  & \texttt{V2X\_Controller}
  & \((5.0,\,1.0,\,0.1)\) & \((1.0,\,0.2,\,0.1)\)
  & \(0.75\) / binary  & \(\pm1.0\)
  & adaptive, cap \(5\mathbf{\text{m/s}}\) \\[2pt]
Perception-swap
  & \texttt{VehiclePIDController}
  & \((5.0,\,1.0,\,0.1)\) & \((0.5,\,0.05,\,0.2)\)
  & \(0.75\) / \(0.5\)  & \(\pm0.8\)
  & \(21.6\,\text{km/h}\) \\
HITL
  & \texttt{VehiclePIDController}
  & \((5.0,\,1.0,\,0.1)\) & \((0.5,\,0.05,\,0.2)\)
  & \(0.75\) / \(0.5\)  & \(\pm0.8\)
  & \(28.8\,\text{km/h}\) \\[2pt]
TCP
  & NN\,+\,PID blend\(^{\ast}\)
  & --- & ---
  & \(0.75\) / cont.  & \(\pm1.0\)
  & emitted by network \\
UniAD
  & \texttt{PIDController\_b2d}
  & \((5.0,\,0.5,\,1.0)\) & \((0.75,\,0.75,\,0.3)\)
  & \(0.75\) / binary  & \(\pm1.0\)
  & adaptive (waypoint-derived) \\
VAD
  & \texttt{PIDController\_b2d}
  & \((5.0,\,0.5,\,1.0)\) & \((0.75,\,0.75,\,0.3)\)
  & \(0.75\) / binary  & \(\pm1.0\)
  & adaptive (waypoint-derived) \\
LMDrive
  & dual PID (LM\(\to\)trajectory)
  & \((5.0,\,0.5,\,1.0)\) & \((1.25,\,0.75,\,0.3)\)
  & \(0.75\) / binary  & \(\pm1.0\)
  & adaptive (waypoint-derived) \\
\bottomrule
\end{tabular}
\end{table*}

\subsection{Evaluation Metrics}

\textbf{Closed-loop Evaluation}. MDrive follows the evaluation protocols of InterDrive \cite{winter2025bevdriver} and the CARLA Leaderboard \cite{carla_leaderboard} and focus on the total end-to-end driving performance. The definition of the closed-loop driving metrics DS and SR is described as following:
\begin{itemize}[leftmargin=*]
    \item \textit{Driving Score (DS)}: The DS measures the overall driving performance, which is the product result of the route completion and the infractions penalty: $DS=RC\times IP$.
    \begin{itemize}
        \item \textit{Route Completion (RC)}: The RC metric is measured by the percentage of the total route distance completed by the testing CAV, which ranges from 0-100\%.
        \item \textit{Infraction Penalty (IP)}: The IP is a safety performance metric that aggregates all unsafe behaviors and traffic violations triggered by the testing CAV during a specific task or route.
    \end{itemize}
    \item \textit{Success Rate (SR)}: The SR represents the robustness and consistency for a driving task, which is the percentage of tasks achieving a full driving score.
\end{itemize}

\textbf{Open-loop Evaluation}. MDrive follows the existing multi-agent open-loop evaluation protocols \cite{lei2025risk,lei2025cooperrisk} to evaluate the perception and planning accuracy with the following metrics.
 \begin{itemize}[leftmargin=*]
    \item \textit{Average Precision (AP)}: The AP evaluates the detection performance of the driving model, typically measured with an Intersection over Union (IoU) threshold of 0.5 for the predicted bounding boxes.
    \item \textit{Average Displacement Error (ADE)}: The ADE measures the planning accuracy by computing the average $L_2$ distance between the planning trajectory of the CAVs and its ground-truth trajectory.
    \item \textit{Collision Rate (CR)}: The CR evaluates the planning safety by calculating the proportion of scenarios where the distance between the CAV planning trajectory and other objects violates a predefined safety threshold. A violation typically occurs if the Time-to-Collision (TTC) is less than 0.9s and the lateral distance is less than 3.5m.
 \end{itemize}

 % We adopt the evaluation protocols of InterDrive \cite{winter2025bevdriver} and the CARLA Leaderboard \cite{carla_leaderboard}, using two metrics for closed-loop driving: \textit{(i) Driving Score (DS)}, which measures overall progress and safety based on route completion and infraction scores; and \textit{(ii) Success Rate (SR)}, the percentage of tasks achieving a full driving score. In open-loop evaluation, we refer to the existing multi-agent open-loop evaluation protocols \cite{lei2025risk,lei2025cooperrisk} to adopt Average Precision (AP) with a union threshold of 0.5 to measure perception accuracy, and leverage Average Displacement Error (ADE) and Collision Rate (CR) to measure the planning safety and accuracy. See~\cref{appendix:mdrive_detailed_settings} for more details.

\section{MDrive Toolbox Details}
\subsection{Additional Details on Human-in-the-Loop Simulation Interface}
\label{subappendix:human_interface}
As \cref{fig:mdrive_toolbox} shows, the core of the Human-in-the-Loop simulation interface is the simulation controller, which collects the human driving input with a gaming steering wheel and pedals. The collected data can be mapped to the same throttle, brake, and steering channels that planners use.

%Talk about collection procedures here
During collections, the simulator runs in synchronous mode with deterministic actor stepping, so a scenario can be paused, rewound, or re-executed deterministically with or without a human in the loop. During takeover, the interface presents a synchronized multi-view overlay: front, side, and rear camera streams from the controlled ego; a top-down bird's-eye view showing all actors, the ego's planned route, and route waypoints; and a screen reporting speed, gear, control inputs, and the current scenario timer. Where relevant, the views can be extended to the other egos' planned trajectories and with the controlled ego's local occupancy, which is useful for occlusion-heavy pre-crash cases where cooperation hinges on what the agent can and cannot see. 

\subsection{Additional Details on Agentic Scenario Generation Pipelines}
\label{subappendix:agentic_pipeine}

% As shown in \textcolor{red}{TODO: agentic workflow chart. The agentic workflow chart should also contain necessary engineering formats, such as prompts, scenario formats, etc., in between.}

\textbf{Agentic Scenario Proposal.} The pipeline begins with an explicit scenario category and runtime configuration, and uses a language agent to generate a constrained JSON schema encoding the intended topology, CAV maneuvers, inter-vehicle constraints, and background actor roles. In this design, the agent serves as a structured scenario proposer rather than a direct generator of simulator assets, making the overall process easier to control, validate, and scale. 

\textbf{Scenario Feasibility Check.} The generated schema is normalized and validated via human-specified rules in the simulator, such as being converted into geometry constraints, matched to compatible map crops, expanded into legal routes, refined into concrete CAV spawns, and populated with scene objects and dynamic actors. Final outputs are exported as scenario directories comprising route waypoint XML files, actor manifests, and optional behavior files. 

\textbf{Duplicate \& Difficulty Screening.} The pipeline also includes a CARLA validation stage to align and repair generated assets before benchmark inclusion. From the raw candidate pool, we curate a final subset of 100 scenarios. We retain 10 scenarios per category through a combination of duplicate filtering, manual review, difficulty screening with a constant-velocity time-to-collision planner, and CARLA-side validation within the generation pipeline. Minor scenario adjustments are applied when needed to improve executability and overall scenario quality. This curation step preserves category balance and removes weak or repetitive variants before inclusion in the benchmark.

\subsection{Additional Details on Real2Sim Pipeline}
\label{subappendix:real2sim_pipeine}
\textbf{Unified Scenario Description}. We construct a unified scenario description~\cite{li2023scenarionet, zhao2026bridgesim} for V2X datasets, and the scenario is represented as a typed record containing (i) an HD map with lane geometry, lane topology, and crosswalks; (ii) a set of actors with per-frame 3D pose, velocity, bounding box, and semantic class (e.g., vehicle, pedestrian, and cyclist); (iii) route and behavior assignment for each CAV and background actors; and (iv) environment objects and the their metadata.

\textbf{Coordinate Alignment and Digital-Twin Mapping.} After we obtain the scenario description, we transform the unified scenario into CARLA's coordinate frame in two stages. First, each real-world driving log is matched to the corresponding real-world HD map and aligned to its digital-twin CARLA counterpart via a global coordinate offset that registers CAV and actor locations to their nominal lanes. For the static environment itself, we follow UrbanVerse~\cite{liu2026urbanverse} and map the real-world scene to CARLA assets, such as lanes, intersections, street furniture, etc., so that the digital twin reflects the structure observed in the source video rather than an arbitrary simulator map. This provides a digital-twin environment in CARLA that reflects real-world scenario layouts and agents behaviors with geometrical and semantic consistency.

\section{More Experimental Details}
The simulation frequency is 20 Hz for closed-loop testing, and the baselines take the last five frames to plan the future trajectory. In the open-loop testing, we evaluate the future three seconds planning trajectory for planning performance.

\subsection{Experimental Detail on \cref{subsec:ol_cl}}
\label{subappendix:ol_cl_exp}

This open-loop to closed loop evaluation gap experiment includes baseline models from all the three model classes in \cref{appendix:simulation_set}  to obtain a comprehensive evaluation comparison.

\textbf{Outlier analysis.} We find that CoDriving and CoLMDriver, though exhibits low perception and prediction results, still exhibit outstanding closed-loop performance. We attribute this as two reasons: 1) The perception mostly misses on far-range perception, where near-range perception are necessary for planning, thus a lack of far-range perceptions might not hurt necessary planning decisions; 2) the other models are more rule-based planning agents connecting to front-end perception results, which might not exhibit outstanding performance as learning-based planners which are trained via high-quality expert demonstrations. These two outliers further explain the discrepancy between open-loop and closed-loop performance, and necessitates the importance of close-loop evaluations on validating methodological developments.

\subsection{Experimental Detail on \cref{subsec:human_study}}
\label{subappendix:human_study}
We designed a human subject research protocol as follows:
\paragraph{Human Participants.} The study recruited three volunteer university students (ages 20–30), all of whom held valid driver's licenses and reported prior video gaming experience. We thoroughly went over the experimental procedures with these participants and let them sign informed consent forms under an IRB-approved protocol. Before the experiment officially started, we explained how predicted trajectories were rendered on screen and allowed each participant to practice with the control interface and learning environments. The study was officially recorded and continued until the participant achieved ten consecutive successful runs.

\paragraph{Main Experiment.} Each participant began with one or two fully manual scenarios to gain familiarity with the tools. Participants were instructed to make decisions at every negotiation step provided by the simulation interface and instructed the decision-making process to maintain safety and follow driving rules. They were directed to prioritize safe task completion and then to guide the agent toward their personal driving preferences.

\section{More Experimental Results}

% \subsection{Model Performance per Interaction Category in MDrive-Interaction}

% \subsection{RQ5: Real-world Deployability}
% We conduct an additional experiment to discuss the real-world deployability of current multi-agent system. Specifically, we are interested in answering \emph{how much does closed-loop performance degrade once system-level latency and inter-agent pose errors are accounted for?}

% To further examine the real-world deployabilities of current multi-agent systems, we follow previous protocols~\cite{zhao2025quantv2x, colmdriver, codriving} examine the closed-loop performance degradations of CoDriving and CoLMDriver brought by 1) system-level latency and 2) pose errors?

\subsection{RQ5: Real-world Deployability}
\label{subsec:real_world_deployability}

We conduct an additional experiment to evaluate the real-world deployability of current multi-agent driving systems. Specifically, we ask: \emph{how much does closed-loop performance degrade once system-level latency and inter-agent pose error are accounted for?} These perturbations are important because V2X cooperation assumes that communicated information is both temporally synchronized and geometrically aligned. In real deployment, however, communication delay, model inference time, and localization uncertainty can break these assumptions.

We evaluate three representative multi-agent systems, CoDriving, CoLMDriver, and CoLMDriver*, on the same matched set of 21 MDrive-V2XPnP scenarios. We report closed-loop Driving Score (DS), Route Completion (RC), Infraction Score (IS), and Success Rate (SR), and compare each perturbed setting against the corresponding clean baseline.

\paragraph{Impact by System-level Latency.}
We first evaluate system-level latency as a temporal perturbation to cooperative driving. In this setting, we introduce a deterministic V2X communication delay of 6 simulation ticks, corresponding to 300 ms at 20 Hz. We also enable real-time execution mode, where the planner, VLM, and LLM-negotiation modules honor their measured wall-clock inference time. During inference, the previous control is replayed until the next action becomes available. Therefore, this setting captures both communication delay and compute-induced action delay, rather than only link latency.

As shown in \cref{tab:latency_results}, all three models experience some degradation under latency, but the degradation pattern is not uniform. CoDriving decreases from 92.85 to 84.59 DS, with SR dropping from 88.10\% to 83.33\%. CoLMDriver is comparatively stable in aggregate, changing only from 79.43 to 78.76 DS, while CoLMDriver* drops from 78.79 to 72.84 DS. These results indicate that latency does not simply reduce performance smoothly across all scenarios. Instead, latency creates sparse but severe closed-loop failures: many scenarios remain largely unchanged, while a small number of latency-sensitive interactions suffer large degradation.

The per-scenario results further support this interpretation. CoDriving maintains strong performance in most cases, but delayed system response causes large failures in specific interaction states where timely recovery is necessary. CoLMDriver* shows similar sensitivity, suggesting that rule-based negotiation can be brittle when timing assumptions are violated. In contrast, CoLMDriver has smaller average degradation, although it still exhibits scenario-level failures. Overall, these results show that strong nominal cooperative performance is not sufficient for robust deployment. Cooperative driving systems must preserve temporal consistency between received information, negotiated decisions, and executed control actions.

\begin{table}[t]
\centering
\caption{\textbf{Impact of system-level latency on MDrive-V2XPnP.} We evaluate three multi-agent models on 21 matched scenarios. The latency setting includes a 6-tick communication delay, corresponding to 300 ms at 20 Hz, together with real-time inference delay.}
\vspace{0.1cm}
\footnotesize
\setlength{\tabcolsep}{3.7pt}
\renewcommand{\arraystretch}{1.12}
\begin{tabular}{l|ccc|ccc|ccc|ccc}
\toprule
\multirow{2}{*}{\textbf{Method}} 
& \multicolumn{3}{c|}{\textbf{DS}$\uparrow$}
& \multicolumn{3}{c|}{\textbf{RC}$\uparrow$}
& \multicolumn{3}{c|}{\textbf{IS}$\uparrow$}
& \multicolumn{3}{c}{\textbf{SR}$\uparrow$} \\
\cmidrule(lr){2-4} \cmidrule(lr){5-7} \cmidrule(lr){8-10} \cmidrule(lr){11-13}
& Clean & Latency & $\Delta$
& Clean & Latency & $\Delta$
& Clean & Latency & $\Delta$
& Clean & Latency & $\Delta$ \\
\midrule
CoDriving & 92.85 & 84.59 & -8.26 & 97.10 & 86.62 & -10.49 & 0.942 & 0.912 & -0.030 & 88.10 & 83.33 & -4.76 \\
CoLMDriver & 79.43 & 78.76 & -0.67 & 82.57 & 81.19 & -1.38 & 0.935 & 0.936 & +0.001 & 71.43 & 69.05 & -2.38 \\
CoLMDriver* & 78.79 & 72.84 & -5.96 & 81.67 & 74.12 & -7.55 & 0.931 & 0.943 & +0.013 & 71.43 & 64.29 & -7.14 \\
\bottomrule
\end{tabular}
\label{tab:latency_results}
\end{table}

\paragraph{Impact by Pose Errors.}
We next evaluate pose error as a spatial perturbation to V2X cooperation. In this experiment, we inject zero-mean Gaussian noise into the cooperator's relative pose, with a position standard deviation of 0.6 m and a rotation standard deviation of 0.6$^\circ$. This perturbation directly tests whether cooperative driving systems remain reliable when shared information is spatially misaligned, which is a realistic failure mode under localization or calibration uncertainty.

As shown in \cref{tab:pose_error_results}, pose error affects the three models unevenly. CoDriving remains nearly unchanged, with DS decreasing only from 92.85 to 92.55 and SR remaining at 88.10\%. This suggests that the perception-sharing policy is relatively robust to moderate localization noise on this matched subset. In contrast, the negotiation-based models degrade more noticeably. CoLMDriver drops from 79.43 to 75.70 DS, while CoLMDriver* drops from 78.79 to 70.63 DS. Both negotiation-based variants also decrease from 71.43\% to 57.14\% SR.

These results indicate that localization error can substantially weaken negotiation-based cooperation. Since negotiation depends on accurate relative geometry among agents, pose misalignment can cause the system to reason over an incorrect interaction state. This can lead to wrong priority assignment, ineffective conflict resolution, and reduced route completion. The larger degradation of CoLMDriver* further suggests that rule-based negotiation is especially sensitive to geometric uncertainty.

Overall, the pose-error experiment shows that robustness to spatial uncertainty is not guaranteed by multi-agent cooperation itself. CoDriving is comparatively stable, while CoLMDriver and especially CoLMDriver* are more sensitive to pose perturbations. This exposes an important deployment gap for negotiation-based cooperative driving: effective communication must be paired with uncertainty-aware reasoning over shared scene geometry.

\begin{table}[t]
\centering
\caption{\textbf{Impact of pose error on MDrive-V2XPnP.} We evaluate three multi-agent models on 21 matched scenarios. The pose-error setting injects zero-mean Gaussian noise into the cooperator's relative pose with $\sigma=0.6$ m for position and $\sigma=0.6^\circ$ for rotation.}
\vspace{0.1cm}
\footnotesize
\setlength{\tabcolsep}{3pt}
\renewcommand{\arraystretch}{1.12}
\begin{tabular}{l|ccc|ccc|ccc|ccc}
\toprule
\multirow{2}{*}{\textbf{Method}} 
& \multicolumn{3}{c|}{\textbf{DS}$\uparrow$}
& \multicolumn{3}{c|}{\textbf{RC}$\uparrow$}
& \multicolumn{3}{c|}{\textbf{IS}$\uparrow$}
& \multicolumn{3}{c}{\textbf{SR}$\uparrow$} \\
\cmidrule(lr){2-4} \cmidrule(lr){5-7} \cmidrule(lr){8-10} \cmidrule(lr){11-13}
& Clean & Pose Err. & $\Delta$
& Clean & Pose Err. & $\Delta$
& Clean & Pose Err. & $\Delta$
& Clean & Pose Err. & $\Delta$ \\
\midrule
CoDriving & 92.85 & 92.55 & -0.30 & 97.10 & 96.28 & -0.82 & 0.942 & 0.942 & +0.000 & 88.10 & 88.10 & +0.00 \\
CoLMDriver & 79.43 & 75.70 & -3.73 & 82.57 & 77.21 & -5.36 & 0.935 & 0.935 & -0.000 & 71.43 & 57.14 & -14.29 \\
CoLMDriver* & 78.79 & 70.63 & -8.16 & 81.67 & 73.10 & -8.57 & 0.931 & 0.934 & +0.004 & 71.43 & 57.14 & -14.29 \\
\bottomrule
\end{tabular}
\label{tab:pose_error_results}
\end{table}

\section{Limitations and Future Work.}
MDrive is built on CARLA and inherits its sensor, dynamics, and rendering fidelity, which still leaves a non-trivial sim-to-real gap relative to on-road deployment. Future work includes: 1) Better integration of cooperative perception and negotiation within a single end-to-end policy, and 2) Closed-loop training and adaptation methods that explicitly target the long-tailed safety-critical interactions where current multi-agent systems still degrade most sharply.

\section{Ethics Statement}
MDrive offers an accessible human-in-the-loop simulation interface that enables human demonstration collections via a user-friendly interface. Since all human studies were carried out in simulation, participants were exposed to no physical risk. Each volunteer provided informed consent, received compensation above prevailing local rates, and was free to pause or exit the study at any moment without any penalty. We capped individual sessions at under one hour and required a minimum three-hour break before a participant could take part again. No personal or sensitive information was gathered or shared, and the project was conducted under IRB approval.

\section{Acknowledgments}
This work was supported by the Federal Highway Administration Center of Excellence on New Mobility and Automated Vehicles, and by the National Science Foundation under Award No. 2346267, POSE: Phase II - DriveX: An Open Source Ecosystem for Automated Driving and Intelligent Transportation Research.

\section{Impact Statement}
\textbf{MDrive Toolbox}. Everyone in the community can leverage the MDrive toolbox to build, extend, and customize their closed-loop cooperative driving scenarios and benchmarks with a low barrier. 

\textbf{MDrive Benchmark}. The MDrive benchmark provides diverse interaction scenarios and Real2Sim scenarios to provide a comprehensive testbed for closed-loop cooperative driving, especially for comparison different cooperative modes, i.e., perception sharing and decision negotiation. 

\textbf{MDrive Challenge}. We are hosting a challenge in the top-tier conference with MDrive Benchmark for advancing methodological developments of multi-agent systems on closed-loop driving. All participants can explore how to leverage V2X communication to facilitate the final planning task, which is a more valuable collaboration paradigm for multi-agent systems, and how to improve the generalization and robustness of cooperative driving within closed-loop evaluation. 

%%%%%%%%%%%%%%%%%%%%%%%%%%%%%%%%%%%%%%%%%%%%%%%%%%%%%%%%%%%%

% \newpage
% \input{checklist.tex}

\end{document}